%% file: preprint-main.tex
\RequirePackage[svgnames]{xcolor}

\documentclass[11pt,letterpaper]{mystyle}
\usepackage[all]{hypcap}
\usepackage[svgnames]{xcolor}
\usepackage[comma,authoryear,compress]{natbib}
\usepackage{hyperref}[citecolor=lightblue]
\usepackage{amsmath}
\usepackage{etoolbox}

\hypersetup{
    colorlinks = true,
    citecolor = {YaleBlue},
}

\usepackage{algorithm}
\usepackage{algorithmicx}
\usepackage{algpseudocode}
\usepackage{microtype}
\usepackage{graphicx}
\expandafter\def\csname ver@subfig.sty\endcsname{}
\usepackage{booktabs} %
\usepackage{float}
\usepackage{bigstrut}

\usepackage{comment}
\usepackage{amsmath}
\usepackage{amssymb}
\usepackage{mathtools}
\usepackage{amsthm}
\usepackage{mathrsfs}
\usepackage{nicefrac}
\usepackage{dsfont}
\usepackage{enumitem}
\usepackage{subcaption}
\usepackage{graphicx,subfig}
\usepackage{cleveref}
\usepackage{bxcoloremoji}
\usepackage{float}
\usepackage{algpseudocode}
\usepackage{xspace}

\usepackage[utf8]{inputenc} %
\usepackage[T1]{fontenc}    %
\usepackage{url}            %
\usepackage{booktabs}       %
\usepackage{amsfonts}       %
\usepackage{nicefrac}       %
\usepackage{microtype}      %
\usepackage{graphicx}
\usepackage{amssymb}
\usepackage{fdsymbol}
\usepackage{wrapfig}
\usepackage{lipsum}
\usepackage{enumitem}
\usepackage{stackengine}
\usepackage[font=small,labelfont=bf]{caption}
\usepackage{color}
\usepackage{adjustbox}

\usepackage{rotating}
\usepackage{makecell}

\usepackage{multirow}

\input{macro}

\newtcolorbox{AIbox}[2][]{aibox,title=#2,#1}
\definecolor{lightblue}{rgb}{0.22,0.45,0.70}%
\definecolor{Gray}{gray}{0.95}
\definecolor{Cornsilk}{rgb}{1.0, 0.97, 0.86}

\usepackage{url}
\usepackage{pifont}
\usepackage{soul}

\usepackage{amsmath}

\usepackage[all]{hypcap}

\usepackage{algpseudocode}
\usepackage{authblk}   
\usepackage{amssymb}      
\usepackage{bbm}          
\usepackage{optidef}  

\usepackage[utf8]{inputenc}
\usepackage[T1]{fontenc}

\usepackage{mathpazo}
\usepackage{tcolorbox}
\tcbuselibrary{skins,breakable,listings} 

\usepackage{lipsum}

\newtcolorbox{simpleElegantQuote}{
    colback=AliceBlue!50!White,   
    colframe=RoyalBlue!75!Black,  
    boxrule=0.5pt,                
    arc=2mm,                     
    boxsep=4pt,                   
    left=10pt, right=10pt,        
    top=8pt, bottom=8pt,         
    fontupper=\itshape,          
}

\input{preamble}

\title{Sandboxed Coding Agents are Competitive Omni-modal Task Solvers}

\runningtitle{}

\author{%
  Dongping Chen\textsuperscript{1*}, Xuanao Huang\textsuperscript{*}, Zhihan Hu, Qingyuan Shi, Dianqi Li, Tianyi Zhou\textsuperscript{2$\ddagger$} \\
  
  \textsuperscript{1}University of Maryland, \textsuperscript{2}MBZUAI
}

\correspondingauthor{Tianyi Zhou: david.tianyi.zhou@gmail.com.}

\begin{document}

\input{Sections/0-abs}

\maketitle

\begin{figure}[!h]
    \centering
    \vspace{-2em}
    \includegraphics[width=0.9\linewidth]{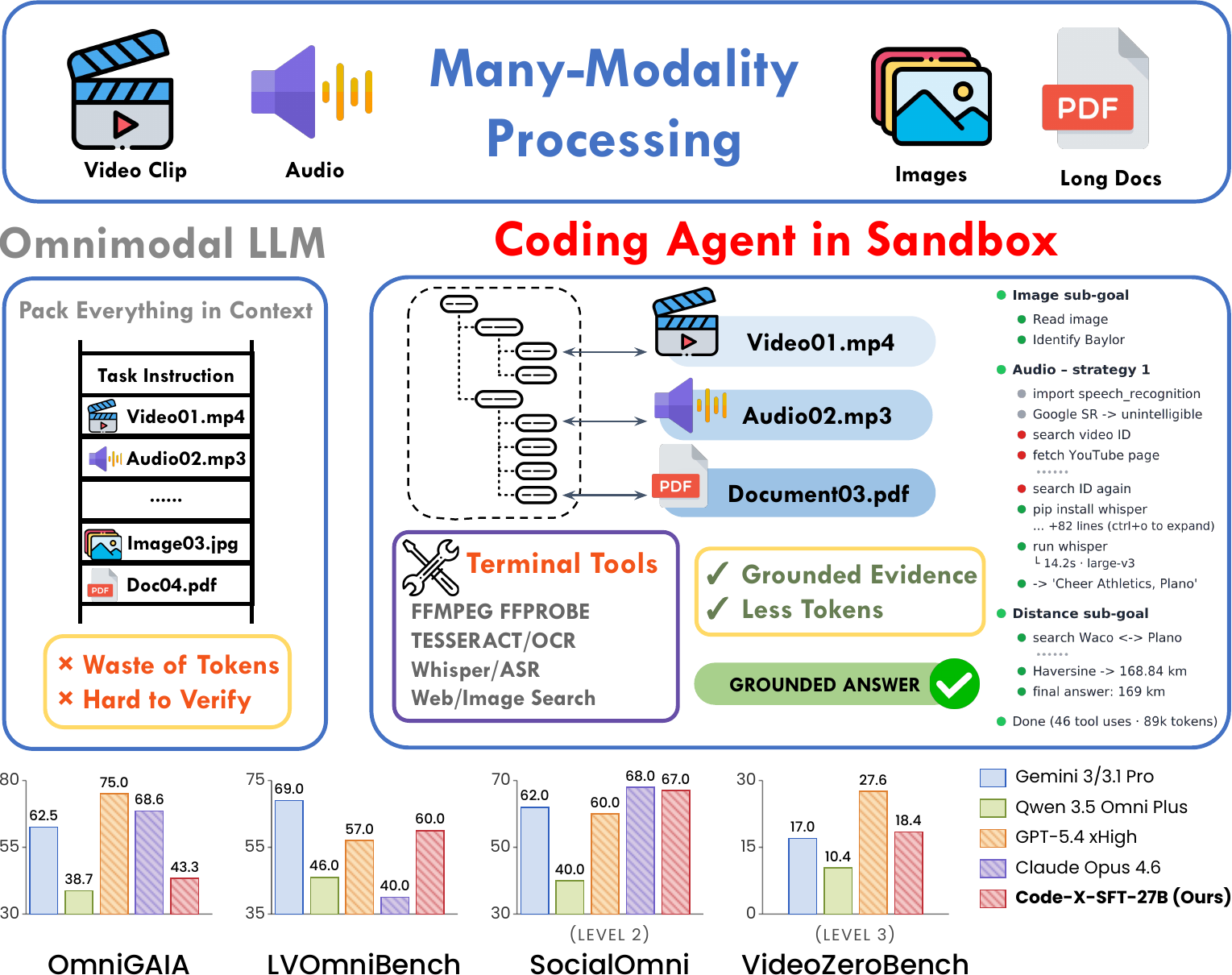}
    \caption{We discover that coding agents are strong omnimodal processors, achieving competitive performance and even surpassing native omnimodal models with fewer tokens on video and audio content through terminal tool-use.}
    \label{fig:new_task}
    \vspace{-1em}
\end{figure}

\input{Sections/1-intro}
\input{Sections/4-analysis}

\input{Sections/5-training}
\input{Sections/6-benchmark}
\input{Sections/2-related}

\input{Sections/10-conclusion}


{
    \small
    \bibliography{custom}
}

\clearpage
\appendix
\input{Sections/99-appendix}

\end{document}

%% file: macro.tex
\definecolor{blanchedalmond}{rgb}{1.0, 0.92, 0.8}
\definecolor{carmine}{rgb}{0.59, 0.0, 0.09}
\definecolor{lightblue}{rgb}{0.22,0.45,0.70}%

\renewcommand{\mathbf}{\boldsymbol}

\makeatletter
\def\Ddots{\mathinner{\mkern1mu\raise\p@
\vbox{\kern7\p@\hbox{.}}\mkern2mu
\raise4\p@\hbox{.}\mkern2mu\raise7\p@\hbox{.}\mkern1mu}}
\makeatother

\definecolor{amaranth}{rgb}{0.9, 0.17, 0.31}
\definecolor{antiquebrass}{rgb}{0.8, 0.58, 0.46}
\definecolor{antiquefuchsia}{rgb}{0.57, 0.36, 0.51}
\definecolor{chromeyellow}{rgb}{0.31, 0.47, 0.26}

%% file: preamble.tex
\usepackage[utf8]{inputenc}
\usepackage[T1]{fontenc}
\usepackage{microtype}

\usepackage{amsmath}
\usepackage{amsfonts}
\usepackage{amssymb}
\usepackage{nicefrac}

\usepackage[dvipsnames,svgnames,x11names,table]{xcolor}

\usepackage{graphicx}
\usepackage{caption}
\usepackage{subcaption}
\usepackage{wrapfig}
\usepackage{placeins}
\usepackage{cuted}
\usepackage{balance}

\usepackage{booktabs}
\usepackage{multirow}
\usepackage{makecell}
\usepackage{tabularx}
\usepackage{array}
\usepackage{adjustbox}
\usepackage{siunitx}
\usepackage{wrapfig}
\usepackage{graphicx}

\usepackage{enumerate}
\usepackage{enumitem}

\usepackage{fancyvrb}
\usepackage{listings}
\usepackage{mdframed}
\usepackage{tcolorbox}

\usepackage{xspace}
\usepackage{soul}
\usepackage{ragged2e}
\usepackage{caption}
\usepackage{float}
\usepackage{pifont}
\usepackage{wasysym}

\definecolor{deeppink}{RGB}{231,84,128}

\usepackage{hyperref}

\newcommand{\header}[1]{\vspace{-0.5em}\smallskip\noindent\textbf{#1}\xspace}

\newcommand{\uniformfont}{\fontsize{9.5pt}{9.5pt}\selectfont}

\newcommand{\method}{\textsc{Code-X}\xspace}
\newcommand{\benchmark}{\textsc{TerminalBench-O}\xspace}
\newcommand{\dataset}{\textsc{OmniCoding}\xspace}

\newcommand{\cmark}{\ding{52}}
\newcommand{\xmark}{\ding{56}}
\newcommand{\halfmark}{\LEFTcircle}

\newcounter{finding}

\newcommand{\finding}[1]{%
  \refstepcounter{finding}%
  \vspace{-0.5em}%
  \begin{tcolorbox}[
    enhanced jigsaw,
    colback=white!90!gray,
    colframe=teal!60!black,
    arc=5pt,
    boxsep=5pt,
    left=1mm,
    right=1mm,
    top=1mm,
    bottom=1mm,
    boxrule=0.8pt,
    drop shadow=gray!50!white
  ]
  \textbf{\textit{Finding \thefinding:}} #1
  \end{tcolorbox}%
}

\newtcbox{\toolcalltag}{
  on line,
  arc=0pt,
  boxrule=0.4pt,
  colback=blue!8,
  colframe=blue!60!black,
  left=2pt,
  right=2pt,
  top=1pt,
  bottom=1pt,
  boxsep=1pt,
  fontupper=\small
}

\newtcbox{\toolresptag}{
  on line,
  arc=0pt,
  boxrule=0.4pt,
  colback=green!8,
  colframe=green!50!black,
  left=2pt,
  right=2pt,
  top=1pt,
  bottom=1pt,
  boxsep=1pt,
  fontupper=\small
}

\newtcbox{\finalanswertag}{
  on line,
  arc=0pt,
  boxrule=0.4pt,
  colback=red!8,
  colframe=red!60!black,
  left=2pt,
  right=2pt,
  top=1pt,
  bottom=1pt,
  boxsep=1pt,
  fontupper=\small
}

\newtcbox{\divergetag}{
  on line,
  arc=0pt,
  boxrule=0.4pt,
  colback=orange!10,
  colframe=orange!70!black,
  left=2pt,
  right=2pt,
  top=1pt,
  bottom=1pt,
  boxsep=1pt,
  fontupper=\small
}

\newcommand{\toolcall}[1]{%
  \par\noindent
  {\toolcalltag{\uniformfont\texttt{tool\_call}}\ #1\ \toolcalltag{\uniformfont\texttt{/tool\_call}}}%
  \par
}

\newcommand{\toolresponse}[1]{%
  \par\noindent
  {\toolresptag{\uniformfont\texttt{tool\_response}}\ #1\ \toolresptag{\uniformfont\texttt{/tool\_response}}}%
  \par
}

\newcommand{\finalanswer}[1]{%
  \par\noindent
  {\finalanswertag{\uniformfont\texttt{final\_answer}}\ #1\ \finalanswertag{\uniformfont\texttt{/final\_answer}}}%
  \par
}

\newcommand{\divergepoint}[1]{%
  \par\noindent
  {\divergetag{\uniformfont\texttt{divergence}}\ #1\ \divergetag{\uniformfont\texttt{/divergence}}}%
  \par
}

\newtcolorbox{sharedprefixbox}{
  enhanced,
  breakable,
  colback=gray!4,
  colframe=black!35,
  boxrule=0.5pt,
  arc=0pt,
  left=4pt,
  right=4pt,
  top=4pt,
  bottom=4pt,
  fontupper=\small,
  fonttitle=\small\bfseries
}

\newtcolorbox{failedtrajbox}{
  enhanced,
  breakable,
  colback=red!2,
  colframe=red!55!black,
  boxrule=0.6pt,
  arc=0pt,
  left=4pt,
  right=4pt,
  top=4pt,
  bottom=4pt,
  title=\textbf{Failed Trajectory},
  colbacktitle=red!10,
  coltitle=black,
  fontupper=\small,
  fonttitle=\small\bfseries
}

\newtcolorbox{successtrajbox}{
  enhanced,
  breakable,
  colback=green!2,
  colframe=green!45!black,
  boxrule=0.6pt,
  arc=0pt,
  left=4pt,
  right=4pt,
  top=4pt,
  bottom=4pt,
  title=\textbf{Successful Trajectory},
  colbacktitle=green!10,
  coltitle=black,
  fontupper=\small,
  fonttitle=\small\bfseries
}

\newtcblisting{agentpromptbox}[2][]{%
  enhanced,
  listing only,
  listing options={
    basicstyle=\ttfamily\scriptsize,
    columns=fullflexible,
    breaklines=true,
    breakatwhitespace=false,
    keepspaces=true,
    showstringspaces=false,
    upquote=true
  },
  attach boxed title to top center={yshift=-3mm,yshifttext=-1mm},
  boxrule=0.9pt,
  colback=gray!00,
  colframe=black!50,
  colbacktitle=gray!25,
  boxed title style={size=small,colframe=gray!50},
  title={#2},
  #1
}

%% file: Sections/0-abs.tex
\begin{abstract}

As multimodal LLMs increasingly target video and audio, it is often assumed that such tasks require native omnimodal models. We show that this is not always the case: coding agents with only text+image access and a sandboxed tool-use interface can match, and in several settings outperform, SOTA native omnimodal models and predefined multimodal agent scaffolds across multiple audio-video benchmarks. Our trajectory analysis suggests that their strength comes from writing code and orchestrating tools to extract relevant evidence from transcripts, frames, and other modality signals, thereby converting omnimodal tasks into retrieval and information-processing problems rather than ingesting entire media streams. We further characterize their limitations through a failure taxonomy and process-level trace analysis, and show that simple skill injection, including human-written and self-distilled skills, substantially improves performance. To explore open-source elicitation, we introduce \method, a training recipe with the \dataset trajectory dataset and verifiable reward, and provide baselines on Qwen-3.5-9B and Qwen-3.6-27B. Finally, we argue that the next frontier is many-modality processing, and introduce \benchmark, a process-level benchmark for real-world omnimodal processing tasks. Code will be available at \href{https://github.com/Dongping-Chen/OmniCoding}{\texttt{https://github.com/Dongping-Chen/OmniCoding}}.

\end{abstract}

%% file: Sections/1-intro.tex
\section{Introduction}
\label{sec:intro}

As AI agents move from screen-bound workflows toward the broader world around users, they are expected to spend increasing amounts of time participating in everyday life as general-purpose assistants. To do so, they must handle multi-modal information from the embodied environments, especially video and audio. This immediately raises a question: Is native video and audio understanding necessary to an agent that is expected to process video or audio in real-world tasks? 

In practice, however, many commonly-adopted agents nowadays, such as OpenClaw \citep{openclaw2026} and Hermes \citep{nousresearch2026hermesagent}, are still text-only or text+image systems, while extending them to handle video and audio natively requires substantial additional modeling and engineering effort. This creates a large gap between real-world agent deployment and the dominant paradigm for video-audio processing, namely native omni-modal models that directly ingest audio and video \citep{qwen35Omni,googleGemini31Pro}. At the same time, modern coding agents already make highly effective use of terminal environments for search, editing, execution, and long-horizon task solving \citep{yang2024sweagent,anthropicClaudeCode,openaiCodexHarness}. This raises a straightforward question: 

\begin{center}
    \textbf{\textit{Can coding agents, which are not designed for native multimodal perception, solve video and audio tasks by exploiting existing tools like \texttt{ffmpeg}, ASR, or new tools they write in a sandbox?}}
\end{center}

We investigate this problem using the most native Codex and Claude Code harnesses, without any human-provided tools or skills. Text+image coding agents, namely GPT-5.4 \texttt{high}/\texttt{xhigh} and Claude Opus 4.6 \citep{openaiGPT54,anthropicOpus46,anthropicClaudeCode}, achieve strong performance on video-audio benchmarks such as OmniGAIA, SocialOmni, and LVOmniBench \citep{li2026omnigaia,xie2026socialomni,tao2026lvomnibench}. Codex with GPT-5.4 \texttt{xhigh} reaches 75.0\% on OmniGAIA without directly perceiving raw video or audio, outperforming the strongest native omni-modal model in our study, Gemini 3.1 Pro \citep{googleGemini31Pro}, by more than 8\%, while Claude Opus 4.6 exceeds it by 2.5\%. Trajectory analysis shows that this advantage does not come from end-to-end perception. Instead, coding agents selectively decompose video and audio into transcripts, sampled frames, and metadata through staged tool use, which also consumes substantially fewer tokens than packing raw streams into model context.

To push coding agents on omni content further, we ask where they still fail and whether such failures can be repaired without retraining the base model. We sample 200 trajectories across models and benchmarks and propose an \textbf{Omni Terminal Failure Taxonomy} together with a process-level evaluation, which reveals six recurring failure modes ranging from premature exploration of multimodal content to brittle tool and environment infrastructure \citep{kim2025omhbench,yang2025avut,liang2025videobrowsecomp}. Guided by this taxonomy, we then study three skill-injection settings inspired by recent work on reusable agent skills and harness optimization \citep{li2026skillsbench,lee2026metaharness}: human-in-the-loop skills that encode expert workflows, calibration-set self-iteration where the agent revises its skills using only binary correctness feedback, and log-driven self-distillation where a separate agent mines reusable skills from raw execution traces. All three improve over the no-skill baseline, raising average OmniGAIA accuracy from 61.4\% to 76.7\% in the best setting, with execution-trace distillation outperforming both binary self-iteration and human-written skills.

The above results are obtained with closed-source frontier models, which raises a natural follow-up question: can the same capability be elicited on open-source models through targeted post-training? As an exploratory baseline, we propose \textbf{\textsc{Code-X}}, a complete recipe for many-modality terminal agents. We construct \textsc{OmniCoding}, a 6{,}035-example trajectory dataset spanning video, audio, image, and cross-modal tasks, and train Qwen-3.5-9B and Qwen-3.6-27B with a supervised warm-start followed by GSPO-based reinforcement learning under a process-aware verifiable reward that jointly penalizes modality-skipping, malformed tool calls, and answer mismatches. Evaluating on four challenging benchmarks, \textsc{Code-X} yields substantial gains over direct MLLM baselines and achieves state-of-the-art results among open-source models, with the 27B model reaching 43.3\% on OmniGAIA, 60.0\% on LVOmniBench, where it surpasses all existing coding agents, and more than doubling performance on VideoZeroBench.

Given the maturity of many-modality understanding, we believe the meaningful frontier lies in many-modality processing: editing videos, auditing image databases, producing spec-compliant multimodal outputs, where coding agents enjoy a structural advantage over end-to-end omni models. To draw attention to this second half of many-modality, we introduce \textbf{\textsc{TerminalBench-O}}, the first process-level benchmark for coding agents on omni-modal processing, with 50 real-world tasks across video, audio, image, document, and cross-modal categories, each evaluated along weighted dimensions over intermediate artifacts and final output compliance. Despite strong omni-understanding results, these tasks remain hard for current agents as they demand long-horizon planning, reliable tool chaining, cross-modal grounding, artifact verification, and strict output adherence; even the strongest GPT-5.5 Codex reaches only 24\% on \textsc{TerminalBench-O}. We hope our analysis, training recipe, and benchmark open new directions for many-modality processing.


%% file: Sections/4-analysis.tex
\vspace{-0.5em}
\section{Analysis of Sandboxed Agent Performance on Omni-Modal Tasks}
\label{sec:analysis}

\header{Preliminary: Adapting Sandboxed Agents for Omni-Modal Tasks.}
We represent each omni-modal task as $x=(q,\mathcal{F},\mathcal{Y})$, where $q$ is a natural-language instruction, $\mathcal{F}$ is a set of input files, and $\mathcal{Y}$ is the set of acceptable final answers (formal definitions and the full rollout formulation are deferred to Appendix~\ref{app:sandbox-formulation}). A native omni-modal model and a sandboxed agent differ mainly in \textit{where} omni-modal processing happens:
\begin{equation*}
\begin{aligned}
\textbf{native omni:}\quad
&(q,\mathcal{F})
\xrightarrow{\mathrm{pack}}
\text{model context}
\xrightarrow{M_\theta}
\hat{y}, \\
\textbf{sandboxed agent:}\quad
&(q,\mathcal{F})
\xrightarrow{\mathrm{stage}}
\text{workspace}
\xrightarrow{\text{terminal interactions}}
\tau
\xrightarrow{\mathrm{Ans}}
\hat{y}.
\end{aligned}
\end{equation*}
A native omni-modal model $M_\theta$ packs all files into model context via media-specific encoding and sampling, treating video and audio as first-class model inputs to generate answers $\hat{y}$. A sandboxed agent, by contrast, stages $\mathcal{F}$ into an isolated workspace and is given only the instruction and file paths; it then inspects files, invokes tools such as \texttt{ffmpeg}, \texttt{ffprobe}, OCR, and ASR, writes and executes scripts, produces intermediate artifacts (transcripts, sampled frames, OCR text, timestamps, tables, crops), which compose a trajectory $\tau$, and submits a final answer $\hat{y}$ parsed from $\tau$. This view treats raw video, audio, images, and documents as environment state rather than mandatory model context, so the agent can solve omni-modal tasks by adaptively reducing raw media into compact evidence that fits the native reasoning interface of a general-purpose coding model.

\header{Agent Harness and Models.}
We use Claude Code \citep{anthropicClaudeCode} and Codex \citep{openaiCodexLoop,openaiCodexHarness} as our two primary agent harnesses. Under CC, we evaluate Claude Opus 4.6 and Claude Sonnet 4.6 \citep{anthropicOpus46,anthropicSonnet46}, as well as MiniMax M2.7 \citep{minimaxM27} and Kimi K2.5 \citep{kimiK25}. Claude Opus 4.6 and Claude Sonnet 4.6 both use the default \texttt{max} setting. Under Codex, we evaluate GPT-5.4 with \texttt{low}, \texttt{medium}, \texttt{high}, and \texttt{xhigh} reasoning effort settings \citep{openaiGPT54}. Throughout evaluation, we follow the official harnesses as closely as possible in order to approximate realistic end-user usage.

For native omni models, we evaluate Gemini 3.1 Pro \citep{googleGemini31Pro}, Gemini 3 Flash \citep{googleGemini3Flash}, Qwen 3.5-Omni \citep{qwen35Omni}, and MiMo-V2-Omni \citep{xiaomiMiMoV2Omni}. We further include two modular omni-agent baselines, OmniAgent \citep{tao2025active} and Agent-Omni \citep{lin2025agentomni}, as strong system-level comparators. Unless otherwise specified, both agent baselines are instantiated with GPT-5.4 using \texttt{high} reasoning effort \citep{openaiGPT54}.

\header{Evaluation Protocol.}
We use four benchmarks with complementary emphases: OmniGAIA for open-ended, tool-mediated factual and multi-hop omni-modal reasoning \citep{li2026omnigaia}, SocialOmni for audio-visual social perception and identity recognition \citep{xie2026socialomni}, LVOmniBench for long-form audio-video understanding \citep{tao2026lvomnibench}, and VideoZeroBench for challenging video-centric reasoning \citep{meng2026videozero}. Given their scale, we evaluate the first 100 examples per level on SocialOmni and LVOmniBench, and on SocialOmni we report only Level-1 and Level-2 Q1 accuracy, omitting the aggregate Level-2 score since it is computed only over Level-2-eligible examples and introduces substantial selection bias. For VideoZeroBench, questions on the same video are asked in a grouped manner. Unless otherwise noted, all models are evaluated with pass@1, and the agent receives only the task instruction and filesystem path(s) to the input files, with no pre-built Skills.

\header{Environments.}
All sandboxed-agent experiments run in a controlled workspace based on \texttt{Ubuntu 22.04.3 LTS} with \texttt{Python 3.13.2} and standard Unix utilities. By default, models may access web search and download additional models or dependencies; we inspect trajectories to detect answer leakage and evaluator hacking. The environment provides common multimedia and scientific tooling (e.g., \texttt{ffmpeg}/\texttt{ffprobe}, Tesseract OCR, ImageMagick, OpenCV, Librosa), and agents may freely install packages and author task-specific scripts for video and audio processing.

\subsection{Main Results and In-depth Analysis}

\begin{table*}[!t]
\centering
\caption{\textbf{Given the limitations of the API and credits before the paper's release, some benchmark results will be updated in the next manuscript.} \textsuperscript{$\ddagger$}Minimax M2.7 uses its own MCP tool for image reading. \textsuperscript{$\S$}Direct image read is disabled for GPT-5.4-\texttt{high} in this ablation setting.}
\label{tab:omni_agent_vs_omnimodal}
\vspace{-0.5em}
\small
\scalebox{0.7}{%
\begin{tabular}{l | c | cccc | cc | cccc | c}
\toprule[1.5pt]
\textbf{\multirow{2}{*}{Model/System}}
& \textbf{\multirow{2}{*}{Harness}}
& \multicolumn{4}{c|}{\textbf{OmniGAIA}}
& \multicolumn{2}{c|}{\textbf{Social Omni}}
& \multicolumn{4}{c|}{\textbf{LVOmniBench}}
& \textbf{VideoZeroBench} \\
\cmidrule(lr){3-6} \cmidrule(lr){7-8} \cmidrule(lr){9-12} \cmidrule(lr){13-13}
&
& Easy & Medium & Hard & Avg.
& Level1 & Level2 Q1 Acc
& Easy & Medium & High & Avg.
& Level 3 \\
\midrule
\multicolumn{13}{c}{\textbf{\textit{Omnimodal LLMs}}} \\
\midrule
Gemini 3.1 Pro
& --
& \textbf{77.8} & \textbf{63.8} & \textbf{52.6} & \textbf{66.1}
& \textbf{90.0} & \underline{62.0}
& \textbf{80.1} & \textbf{71.8} & \textbf{48.0} & \textbf{69.0}
& \underline{17.6} \\
Gemini 3 Flash
& --
& \underline{67.2} & \underline{46.9} & \underline{37.2} & \underline{51.7}
& 85.0 & \textbf{67.0}
& \underline{72.2} & \underline{64.1} & 32.0 & \underline{59.0}
& \textbf{17.8} \\
Qwen 3 Omni
& --
& 19.7 & 10.6 & 9.0 & 13.3
& 61.0 & 39.0
& 30.6 & 17.9 & 28.0 & 25.0
& 5.2 \\
Mimo V2 Omni
& --
& 33.9 & 22.5 & 19.1 & 25.8
& 64.0 & 40.0
& 51.4 & 25.2 & 24.0 & 35.0
& 7.4 \\
Qwen 3.5 Omni Plus
& --
& 43.5 & 39.4 & 29.0 & 38.7
& \underline{88.0} & 40.0
& 58.3 & 35.9 & \underline{44.0} & 46.0
& 10.4 \\
\midrule
\multicolumn{13}{c}{\textbf{\textit{Coding Agents (Our Findings)}}} \\
\midrule
GPT-5.4 Low
& Codex
& 63.1 & 53.8 & 44.9 & 55.0
& 64.0 & 59.0
& 58.3 & 46.2 & 36.0 & 48.0
& 21.0 \\
GPT-5.4 Medium
& Codex
& 66.4 & 55.0 & 43.6 & 56.4
& 71.0 & 49.0
& \textbf{72.2} & 43.6 & 40.0 & 53.0
& 24.0 \\
GPT-5.4 High
& Codex
& 70.4 & 60.0 & 50.0 & 61.4
& \underline{72.0} & 51.0
& 66.6 & \underline{59.0} & \textbf{44.0} & \textbf{58.0}
& \underline{27.2} \\
GPT-5.4 xHigh
& Codex
& \textbf{82.0} & \textbf{75.0} & \textbf{64.1} & \textbf{75.0}
& \textbf{75.0} & \underline{60.0}
& \underline{66.7} & \textbf{64.1} & 32.0 & \underline{57.0}
& \textbf{27.6} \\
Claude Opus 4.6
& CC
& 74.6 & \underline{69.4} & \underline{57.7} & \underline{68.6}
& 54.0 & \textbf{68.0}
& 47.2 & 41.0 & 28.0 & 40.0
& - \\
Claude Sonnet 4.6
& CC
& 73.8 & 62.5 & 48.8 & 63.3
& - & -
& - & - & - & -
& - \\
MiniMax-M2.7\textsuperscript{$\ddagger$}
& CC
& 38.5 & 34.4 & 23.1 & 33.3
& 28.0 & 46.0
& 50.0 & 15.4 & 24.0 & 30.0
& 5.8 \\
Kimi K-2.5
& CC
& 72.1 & 50.6 & 39.7 & 55.6
& 46.0 & 50.0
& 55.6 & 28.0 & \underline{41.0} & 43.0
& 15.6 \\
\midrule
\multicolumn{13}{c}{\textbf{\textit{Tool-use Agent (GPT-5.4 High as backbone)}}} \\
\midrule
OmniAgent
& -
& \underline{77.9} & 63.1 & 55.1 & 66.4
& 55.0 & 56.0
& 50.0 & 21.0 & 28.0 & 33.0
& 12.2 \\
Agent-Omni
& -
& 54.1 & 41.9 & 19.2 & 41.1
& 37.0 & 48.0
& 56.0 & 28.0 & 20.0 & 36.0
& 5.0 \\
\midrule
\multicolumn{13}{c}{\textbf{\textit{Text-Only Coding Agent}}} \\
\midrule
GPT-5.4 High\textsuperscript{$\S$}
& Codex
& 70.5 & 63.8 & 56.4 & 64.4
& 74.0 & 48.0
& 52.8 & 38.5 & 28.0 & 41.0
& 19.0 \\
\bottomrule[1.5pt]
\end{tabular}
}

\end{table*}

From Table~\ref{tab:omni_agent_vs_omnimodal}, coding agents prove highly competitive with native omnimodal LLMs across all four benchmark suites. On OmniGAIA, GPT-5.4 \texttt{xhigh} (75.0) clearly outperforms Gemini 3.1 Pro (66.1), especially on medium- and hard-level tasks; on VideoZeroBench, the best coding agent (27.6) also exceeds Gemini 3 Flash (17.8). Native omni models lead on Social Omni Level-1, though Claude Opus 4.6 slightly edges out the best native score on Level-2 Q1. On LVOmniBench, Gemini 3.1 Pro remains the strongest overall, but coding agents stay competitive on the easy and medium subsets. These patterns suggest that benchmarks emphasizing fine-grained temporal perception and audio--visual alignment favor end-to-end omni modeling, while tool-oriented benchmarks such as OmniGAIA align more naturally with the coding-agent paradigm.
\finding{Coding agents with only text+image capability can perform competitively with, and in some cases outperform, \emph{state-of-the-art} native omnimodal LLMs on omnimodal tasks.}

Across the GPT-5.4 variants under the same Codex harness, increased reasoning effort yields clear gains on omnimodal performance. On OmniGAIA, average accuracy rises steadily from 55.0 (\texttt{low}) to 56.4 (\texttt{medium}), 61.4 (\texttt{high}), and 75.0 (\texttt{xhigh}), with improvements appearing across easy, medium, and hard subsets. The same trend is visible on LVOmniBench, where higher-effort settings outperform lower-effort ones, though not strictly monotonically. Since the model family and harness are held fixed, this isolates the contribution of inference-time computation alone. 
\finding{Increasing reasoning effort generally improves coding-agent performance, suggesting that omnimodal task success depends not only on model perception capacity but also on the depth of agentic computation.}


\begin{figure*}[!t]                                            
    \centering     
    \includegraphics[width=\linewidth]{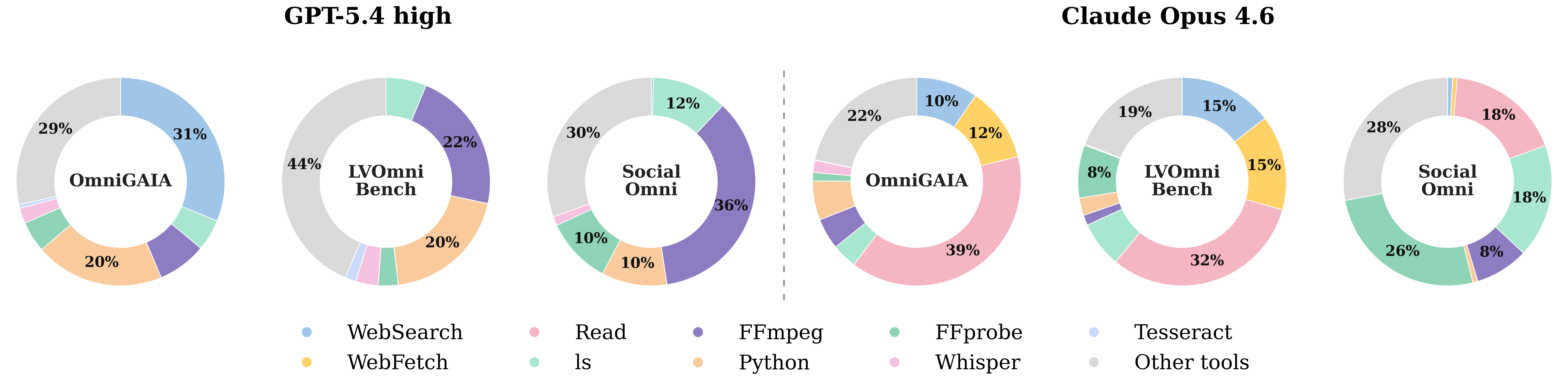}
    \vspace{-2em}
    \caption{Tool-use distributions of GPT-5.4 \texttt{high} and Claude Opus 4.6 \texttt{max} across four benchmarks. }  
    \label{fig:tool_use}  
    \vspace{-1em}                                      
\end{figure*}


\header{Tool-use Analysis.} We analyze tool-use behavior in coding agents. Table~\ref{tab:tool_use_merged} report the six most behaviorally salient tools (\texttt{ffmpeg}, \texttt{ffprobe}, \texttt{python3}, \texttt{whisper}, \texttt{WebSearch}, \texttt{tesseract}), together with their average calls per case (\textit{Avg.}), success/failure gap (\textit{Gap}, in percentage points), and total tool calls per case (\textit{Avg. Tool Use}); Figure~\ref{fig:tool_use} further visualizes overall usage patterns for GPT-5.4 \texttt{high} and Claude Opus 4.6 \texttt{Max}. Across benchmarks, coding agents handle non-text inputs via a staged, tool-mediated pipeline in which \texttt{ffprobe}/\texttt{ffmpeg} inspect and transform media while \texttt{whisper} and \texttt{python3} support transcription and downstream analysis. Media extraction, transcription, and external search yield the clearest positive gains, whereas \texttt{python3}, our proxy for self-authored tools, is frequently invoked but shows mixed gains, suggesting ad-hoc scripting is common but not reliably beneficial. Holding the harness and model family fixed, higher reasoning-effort settings tend to use more tools, most visibly on OmniGAIA and LVOmniBench.
 
\finding{Omnimodal problem solving proceeds through a staged tool-mediated pipeline, where media extraction, transcription, and external search (\texttt{ffmpeg}, \texttt{ffprobe}, \texttt{whisper}, \texttt{web\_search}) deliver the most consistent gains. Agents invoke \texttt{python3} frequently but unreliably; tool-use performance scales with reasoning effort when harness and model family are fixed.}

\begin{figure}[!t]
    \centering
    \includegraphics[width=0.9\linewidth]{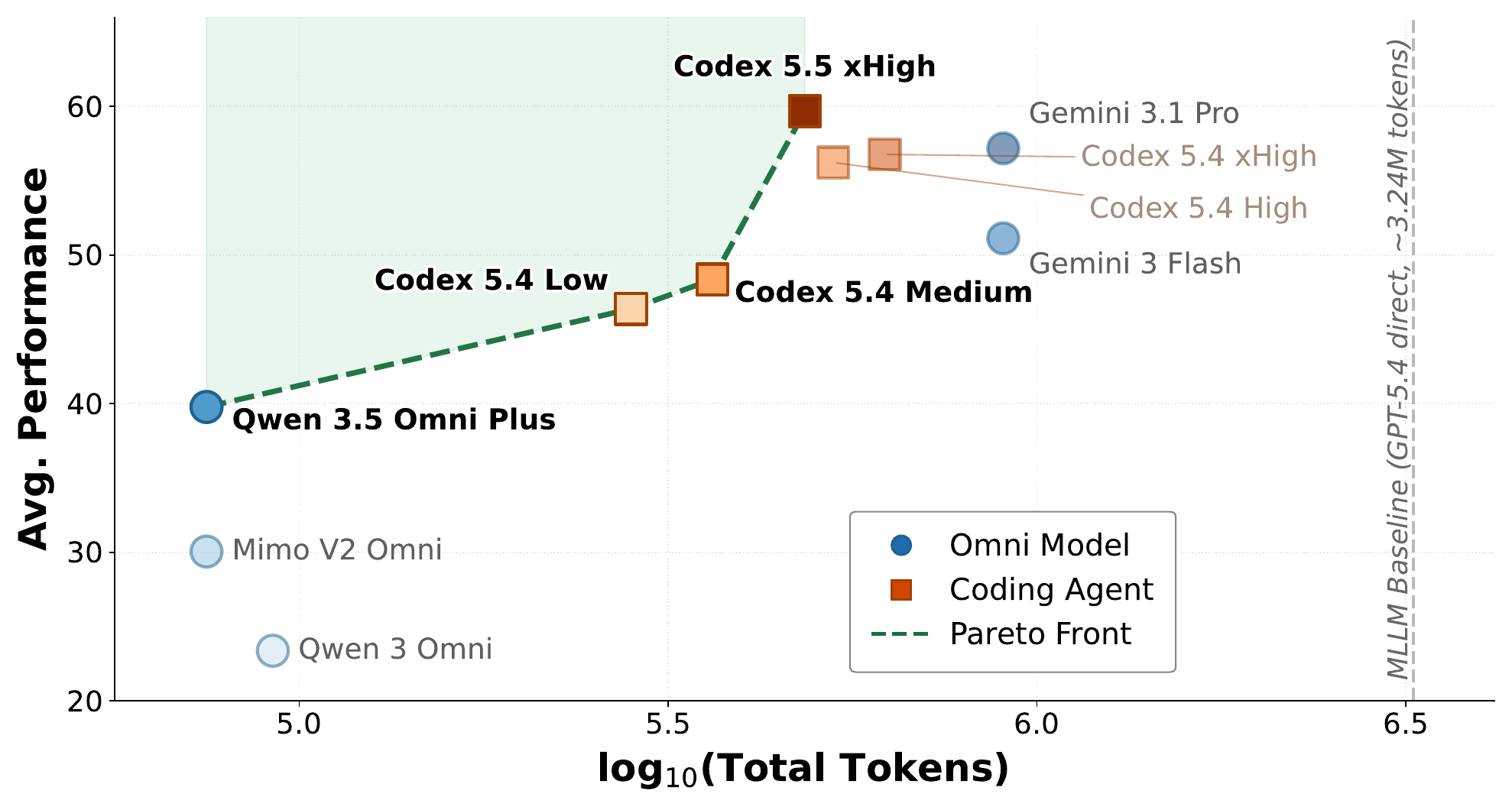}
    \vspace{-0.5em}
    \caption{Pareto-front of Acc-Token tradeoff. We find that sandboxed coding agents are efficient and competitive omnimodal task solvers on avg. of four benchmarks. The gray line indicates the estimated MLLM baseline. }
    \label{fig:token_usage}
    \vspace{-1em}
\end{figure}

\header{Token Usage Analysis.} Given the concern of using more tokens for better performance, we calculate the avg. token input and accuracy across four benchmarks. As shown in Figure~\ref{fig:token_usage}, surprisingly, coding agents consume much fewer tokens compared to omnimodal models. Based on our analysis of model trajectories and tool use patterns, we found that the image and audio content actually consumed by the coding agent is far smaller than what would otherwise be fed in its entirety to an omnimodal model. Through proactive tool use, the model selectively retrieves only the content it needs, substantially reducing the multimedia footprint in its context and thereby eliminating a significant amount of redundant tokens.

\header{Text-Only Ablation.}
Since some agents (\emph{e.g.}, DeepSeek V4 \citep{deepseekai2026deepseekv4}, MiniMax M2.7 \citep{minimaxM27}) are text-only LLMs, we include a text-only ablation in Table~\ref{tab:omni_agent_vs_omnimodal}, evaluating Codex GPT-5.4-\texttt{high} under the same coding-agent setup but without direct image access. It remains competitive with many native omnimodal LLMs, showing only a modest gap to its vision-enabled counterpart on \textsc{OmniGAIA} and \textsc{SocialOmni} even matching or slightly exceeding it, while on the more vision-centric \textsc{LVOmniBench} and \textsc{VideoZeroBench} performance drops sharply. This exposes a limitation of current omnimodal benchmarks: many tasks can be solved through text or shallow perception (e.g., OCR) rather than deep visual understanding.

\label{sec:failure-taxonomy}

\begin{figure*}[!t]
    \centering
    \includegraphics[width=\linewidth]{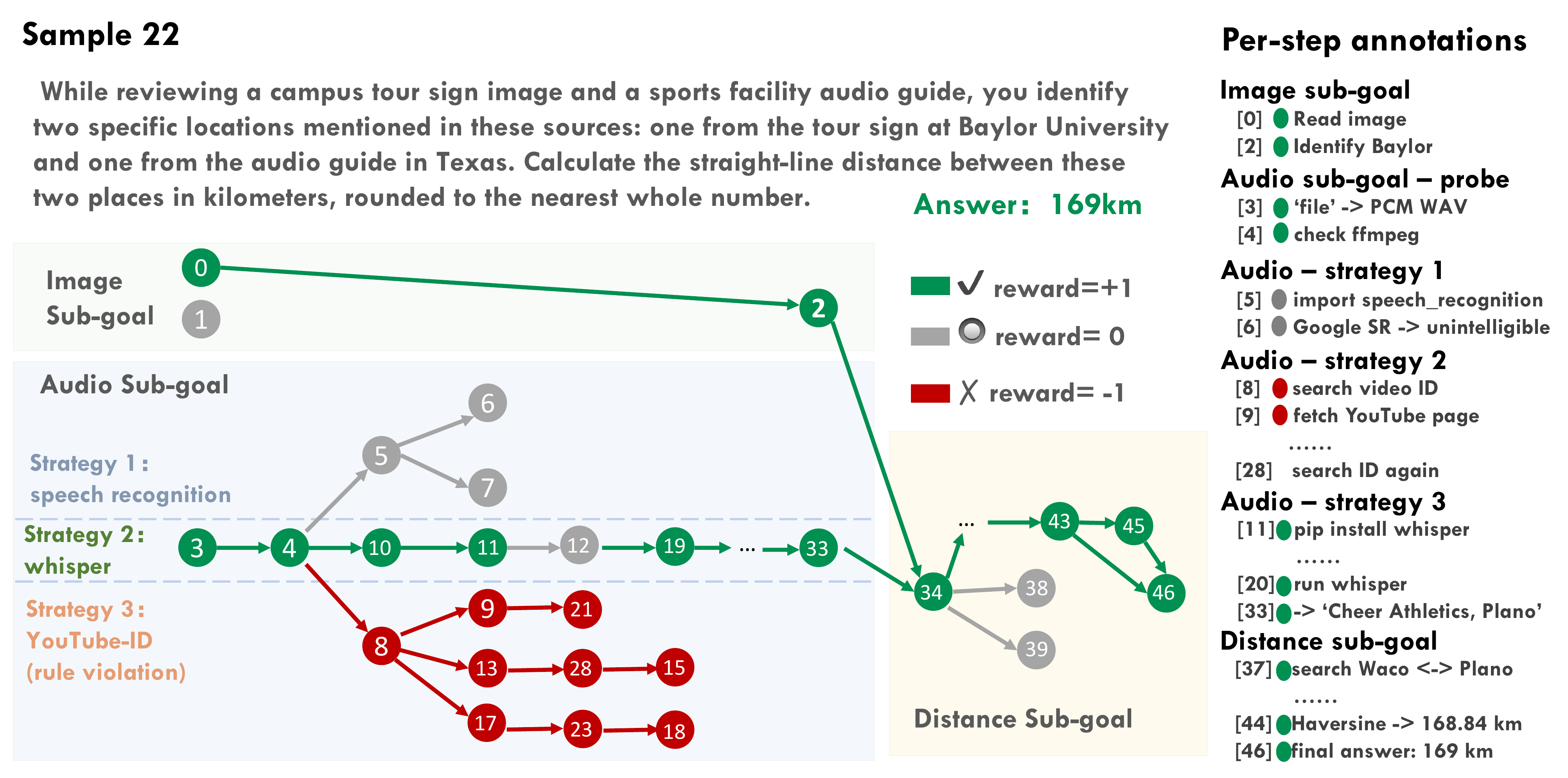}
    \vspace{-1em}
    \caption{Trajectory DAG of a \textsc{OmniGAIA} sample (Baylor campus-tour sign $\times$ Texas sports-facility audio; ground-truth $169$~km). Nodes are annotated as agent steps, coloured by the step-supervisor reward ($\checkmark$ / $0$ / $\times$ for $+1$ / $0$ / $-1$). \emph{Audio} sub-goal includes three parallel strategies: \emph{speech\_recognition}, \emph{whisper}, and \emph{YouTube-ID lookup}; evidence from the \emph{Image} and \emph{Audio} sub-goals merges into \emph{Distance}. }
    \label{fig:trajectory_dag_case}
    \vspace{-1em}
\end{figure*}


\begin{figure*}[!h]
    \centering
    \includegraphics[width=0.95\linewidth]{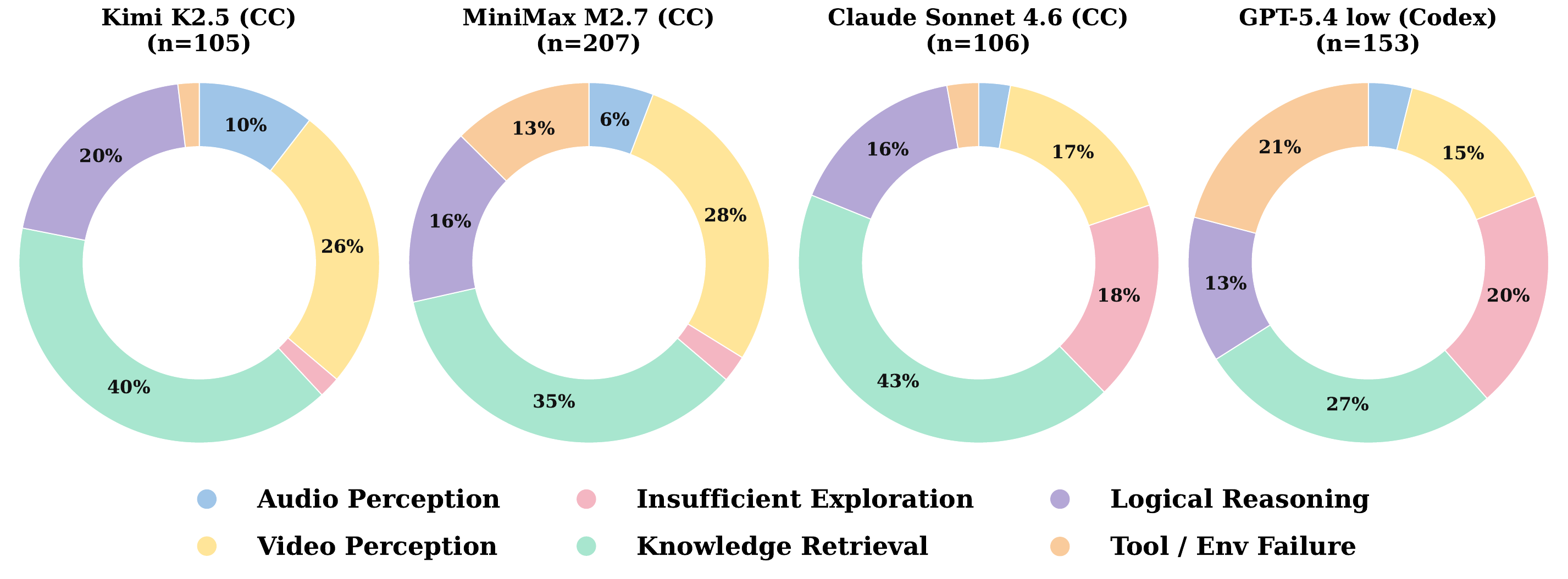}
    \vspace{-0.5em}
    \caption{Distribution of primary error types across four coding agents on \textsc{OmniGAIA}, one pie per agent.  The sample size $n$ is shown under each agent name. }
    \label{fig:error_types_by_model}
    \vspace{-1em}
\end{figure*}

\begin{table*}[!h]
\centering
\caption{Tool-use statistics of representative multimodal tools across coding-agent systems on \textit{OmniGAIA}, \textit{SocialOmni} (Level 1 \& Level 2), and \textit{LVOmniBench}. ``Avg.'' denotes the average number of calls per case. ``Gap'' is reported in percentage points and denotes \textit{$\text{success\_case\_ratio} - \text{failure\_case\_ratio}$}. Positive gaps indicate tools more commonly used in successful cases, while negative gaps indicate tools more associated with failure.}
\label{tab:tool_use_merged}
\vspace{-0.5em}
\small
\scalebox{0.7}{%
\begin{tabular}{l | c | c | cc | cc | cc | cc | cc | cc}
\toprule[1.5pt]
\multirow{2}{*}{\textbf{System}}
& \multirow{2}{*}{\textbf{Harness}}
& \multirow{2}{*}{\textbf{Avg. Tool Use}}
& \multicolumn{2}{c|}{\textbf{ffmpeg}}
& \multicolumn{2}{c|}{\textbf{ffprobe}}
& \multicolumn{2}{c|}{\textbf{python3}}
& \multicolumn{2}{c|}{\textbf{whisper}}
& \multicolumn{2}{c|}{\textbf{web\_search}}
& \multicolumn{2}{c}{\textbf{tesseract}} \\
\cmidrule(lr){4-5} \cmidrule(lr){6-7} \cmidrule(lr){8-9} \cmidrule(lr){10-11}
\cmidrule(lr){12-13} \cmidrule(lr){14-15}
& &
& Avg. & Gap (\%)
& Avg. & Gap (\%)
& Avg. & Gap (\%)
& Avg. & Gap (\%)
& Avg. & Gap (\%)
& Avg. & Gap (\%) \\
\midrule
\multicolumn{15}{c}{\textbf{\textit{OmniGAIA}}} \\
\midrule
Kimi K-2.5         & CC    & 22.63 & 1.45 & +12.1          & 0.34 & +7.5          & 2.15 & +9.1           & 0.62 & +6.6           & 2.89 & \textbf{+19.2} & 0.44 & +0.0 \\
MiniMax-M2.7       & CC    & 44.86 & 1.50 & -0.8           & 0.49 & -8.3          & 7.43 & -5.4           & 0.31 & +0.8           & 2.94 & \textbf{+2.9}  & 0.06 & -0.8 \\
Claude Sonnet 4.6  & CC    & 40.85 & 1.97 & +0.4           & 0.54 & +1.2          & 2.49 & \textbf{+16.5} & 0.77 & +8.9           & 3.90 & +14.5          & 0.02 & +0.2 \\
GPT-5.4 Low        & Codex & 25.50 & 2.59 & -2.3           & 1.19 & -6.6          & 4.36 & -9.9           & 0.08 & \textbf{+2.7}  & 5.74 & +1.4           & 0.69 & -9.9 \\
GPT-5.4 Medium     & Codex & 31.83 & 3.05 & -4.0           & 1.25 & -0.1          & 4.98 & -6.6           & 0.06 & \textbf{+1.5}  & 7.97 & -4.3           & 0.81 & -7.3 \\
\midrule
\multicolumn{15}{c}{\textbf{\textit{SocialOmni Level1}}} \\
\midrule
GPT-5.4 Low        & Codex & 16.28 & 6.94 & \textbf{+11.1} & 2.04 & +4.0           & 2.94  & +9.4          & 0.22 & -1.7           & 0.00 & +0.0           & 0.00 & +0.0  \\
GPT-5.4 Medium     & Codex & 21.52 & 9.18 & +3.4           & 2.06 & +0.0           & 3.22  & +3.2          & 0.52 & \textbf{+24.7} & 0.00 & +0.0           & 0.00 & +0.0  \\
GPT-5.4 High       & Codex & 19.68 & 5.06 & +12.9          & 2.18 & \textbf{+16.5} & 2.32  & +10.6         & 0.46 & +12.2          & 0.00 & +0.0           & 0.00 & +0.0  \\
GPT-5.4 xHigh      & Codex & 39.90 & 9.24 & +33.3          & 2.06 & \textbf{+34.7} & 15.64 & +28.0         & 1.22 & +14.7          & 0.00 & +0.0           & 0.04 & -2.7  \\
\midrule
\multicolumn{15}{c}{\textbf{\textit{SocialOmni Level2}}} \\
\midrule
GPT-5.4 Low        & Codex & 5.59  & 1.98 & -4.0           & 0.76 & -5.2           & 0.41 & -5.7           & 0.01 & -2.4           & 0.00 & \textbf{+0.0}  & 0.00 & \textbf{+0.0} \\
GPT-5.4 Medium     & Codex & 10.07 & 4.92 & +0.0           & 1.20 & \textbf{+2.0}  & 1.28 & -16.8          & 0.04 & -1.9           & 0.00 & +0.0           & 0.00 & +0.0          \\
GPT-5.4 High       & Codex & 13.47 & 6.75 & +0.0           & 1.25 & +0.0           & 1.05 & -3.1           & 0.00 & +0.0           & 0.10 & \textbf{+5.9}  & 0.00 & +0.0          \\
GPT-5.4 xHigh      & Codex & 12.76 & 3.67 & -0.8           & 1.26 & +0.8           & 1.55 & -9.2           & 0.17 & \textbf{+6.7}  & 0.00 & +0.0           & 0.00 & +0.0          \\
\midrule
\multicolumn{15}{c}{\textbf{\textit{LVOmniBench}}} \\
\midrule
GPT-5.4 Low        & Codex & 31.20 & 7.10  & \textbf{+0.0} & 1.40 & \textbf{+0.0}  & 6.45 & -16.0          & 1.31 & -9.5           & 0.00 & \textbf{+0.0}  & 0.44 & -8.0          \\
GPT-5.4 Medium     & Codex & 28.72 & 6.62  & +2.1          & 1.26 & +0.0           & 4.67 & +12.8          & 1.46 & \textbf{+15.9} & 0.00 & +0.0           & 0.29 & +8.2          \\
GPT-5.4 High       & Codex & 38.75 & 8.56  & -3.4          & 1.20 & +0.0           & 7.64 & -15.0          & 1.34 & -1.6           & 0.00 & +0.0           & 0.66 & \textbf{+3.8} \\
GPT-5.4 xHigh      & Codex & 42.29 & 10.93 & +0.6          & 1.23 & +0.0           & 8.21 & -4.2           & 1.65 & \textbf{+13.2} & 0.00 & +0.0           & 0.69 & +10.1         \\
\bottomrule[1.5pt]
\end{tabular}
}
\vspace{-1em}
\end{table*}

\subsection{Failure Analysis: Taxonomy and Process-level Trajectory.} 
Given the strong performance of coding agents, we investigate the challenges they face in omni content processing. We propose a new failure mode taxonomy based on task type, and sample 200 trajectories (6,268 steps in total) across different models and benchmarks, along with their primary failure reasons. For each trajectory, we construct a DAG to identify its main trunk and branches, enabling process-level annotation. The resulting taxonomy comprises six failure modes: (i) \textbf{Audio Perception and Extraction}, (ii) \textbf{Video Perception and Extraction}, (iii) \textbf{Insufficient Exploration of Modal Content}, (iv) \textbf{Knowledge Retrieval and Factual}, (v) \textbf{Logical Reasoning and Calculation}, and (vi) \textbf{Tool and Environment Infrastructure}. Detailed definitions and representative examples are deferred to Appendix~\ref{appendix:failure_modes}. To scale the analysis beyond the 200 hand-annotated trajectories, we use GPT-5.5 as the process-level judge, which we validate against human labels on this small benchmark (see Appendix~\ref{appendix:llm_judge} for the validation protocol and results).

Figure~\ref{fig:error_types_by_model} reports the empirical distribution of the six failure modes across four coding agents on \textsc{OmniGAIA}. Knowledge-Retrieval dominates the residual errors for three of four agents (Kimi~40\%, Sonnet~43\%, MiniMax~35\%), while Codex-GPT5.4-Low is distinctive in that Tool/Environment Failure (21\%) and Insufficient Exploration (20\%) each rival its knowledge errors (27\%). MiniMax further stands out for heavy Video-Perception errors (28\%), indicating that different agents break down along different axes of the taxonomy. Beyond aggregate rates, Figure~\ref{fig:trajectory_dag_case} dissects a single \textsc{OmniGAIA} trajectory as a DAG of annotated steps, showing how multiple modes from our taxonomy---audio-perception errors, rule violations, and scraping failures---co-occur with exploration-with-recovery behavior inside the same run, and how the agent ultimately recovers the correct answer.

\input{Sections/case-study-figure}

\subsection{Skills and Self-Evolution for Strong Many-modality Content Processing}
\header{Experiment Setups.}
Following recent work, we treat Skills as structured, inference-time packages of procedural knowledge that supply reusable workflows, instructions, and optionally executable resources to an agent \citep{li2026skillsbench,xu2026agentskills}. Unless explicitly stated otherwise, the main results in this paper are reported without Skills.
We study three settings as follows:
\begin{itemize}[leftmargin=*,itemsep=0pt]
    \item \textbf{Calibration-set self-iteration.} The agent is given a small calibration set and iteratively revises its Skills using only binary correctness feedback; we reveal whether an attempt is correct or incorrect, but do not provide any human-written diagnosis of the intermediate process.
    \item \textbf{Log-driven self-distillation.} A separate agent reviews raw execution logs, extracts recurring success and failure patterns, and writes Skills from these traces. This setting is inspired by recent log- and trace-driven harness optimization work such as Meta-Harness \citep{lee2026metaharness}.
    \item \textbf{Human-in-the-loop Skills.} We provide manually written Skills that encode a strong human route or best-known workflow. This setting serves as an approximate upper bound on the value of explicit procedural knowledge, while also introducing human prior and route bias.
\end{itemize}

\begin{wrapfigure}{r}{0.49\linewidth}
    \centering
    \vspace{-1em}
    \includegraphics[width=\linewidth]{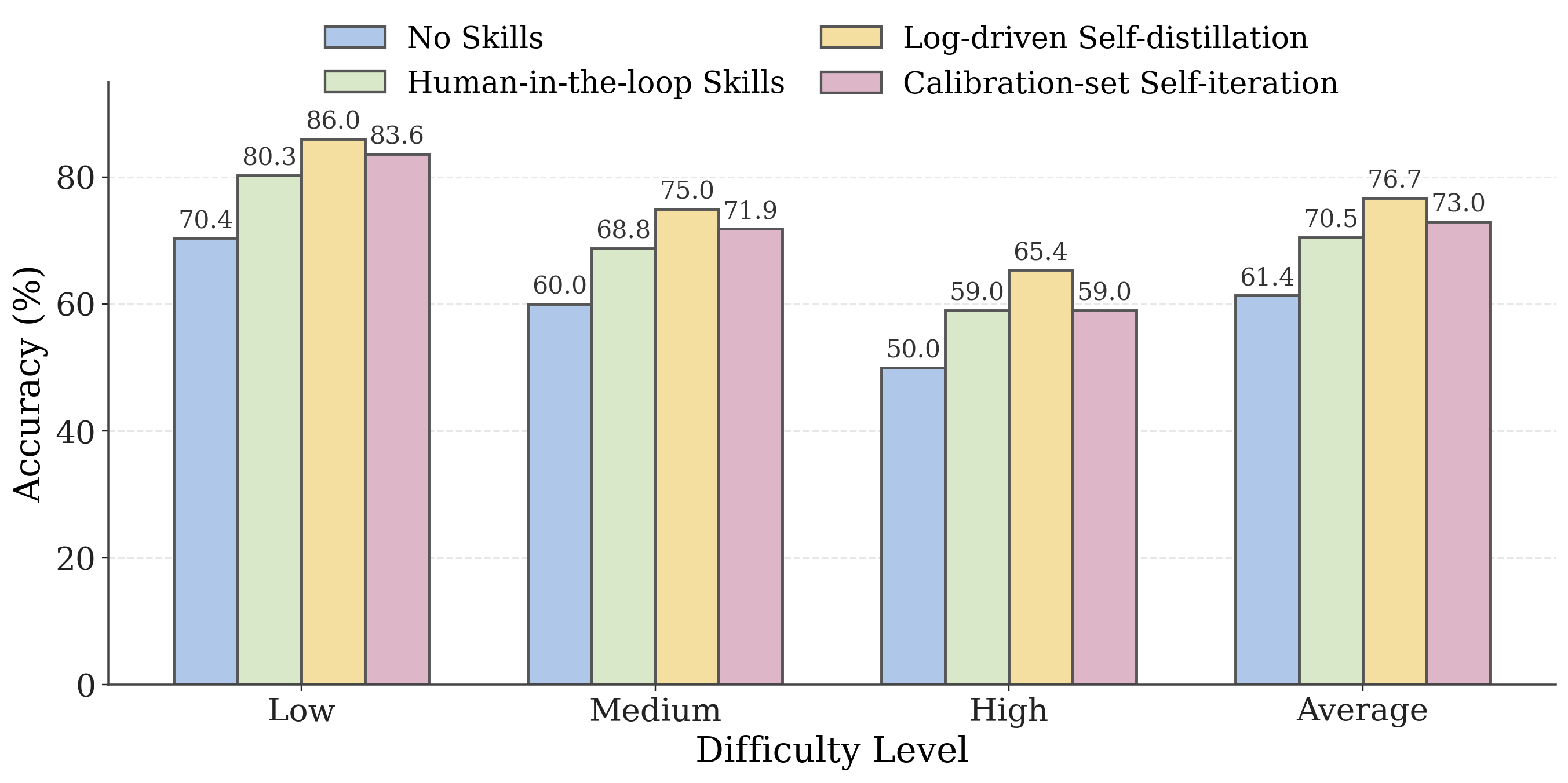}
    \vspace{-1.5em}
    \caption{Performance of GPT-5.4 \texttt{high} on OmniGAIA under four settings.}
    \label{fig:omnigaia-skills}
    \vspace{-1em}
\end{wrapfigure}

\header{Results and Analysis.}
Figure~\ref{fig:omnigaia-skills} shows that explicit \textit{Skills} consistently improve \textsc{OmniGAIA} performance over the \textit{No Skills} baseline across all difficulty levels. All skill-enhanced variants raise average accuracy, with gains remaining visible on the \textit{High} split. The best setting improves average accuracy from 61.4\% to 76.7\%, suggesting that inference-time procedural knowledge can substantially strengthen omni-modal agent performance. Figure~\ref{fig:idx161_tofino_case} provides a representative example, where Skills help the agent avoid an incorrect application of regional rules and reach the correct answer.

Both self-improving methods outperform the baseline, with \textit{Log-driven self-distillation} achieving the strongest results: 76.7\% average accuracy versus 73.0\% for \textit{Calibration-set self-iteration}, and the gap widens on harder problems (65.4\% vs.\ 59.0\% on the \textit{High} split). Execution traces appear to provide richer supervision than binary feedback from a small calibration set, aggregating recurring success and failure patterns from broader agent behavior rather than adapting to local calibration-set patterns, likely because they offer process-level signals with broader coverage, enabling more stable and generalizable skill refinement in practice.

\finding{Execution-trace-based skill refinement yields more effective and robust improvements than both calibration-based self-iteration and manually written Skills.}

%% file: Sections/case-study-figure.tex

\begin{figure}[!h]
\centering
\begingroup
\uniformfont
\setlength{\parskip}{0.25em}
\setlength{\parindent}{0pt}

\begin{tcolorbox}[enhanced,
  attach boxed title to top center={yshift=-2.5mm,yshifttext=-0.8mm},
  boxrule=0.7pt,
  colback=gray!00,
  colframe=black!50,
  colbacktitle=Gray,
  title=Case Study,
  fonttitle=\uniformfont\bfseries,
  fontupper=\uniformfont,
  boxed title style={
    size=small,
    colframe=Gray,
    boxrule=0.5pt,
    boxsep=0.4mm,
    left=1mm,
    right=1mm,
    top=0.4mm,
    bottom=0.4mm
  },
  boxsep=0.7mm,
  left=1.1mm,
  right=1.1mm,
  top=1.0mm,
  bottom=1.0mm,
]

\uniformfont

\noindent\textbf{Question.}
Watch the video showcasing a prominent surfing destination in British Columbia, Canada. Identify the specific town featured. Then, according to the Canadian Marine Mammal Regulations and 2024/2025 management measures for killer whale protection, what is the minimum mandatory vessel approach distance in meters required for killer whales in the waters immediately offshore of that town?

\noindent\textbf{Labeled Answer.}
200

\tcblower

\noindent\textbf{Shared Prefix.}

\begin{sharedprefixbox}
\uniformfont
\setlength{\parskip}{0.25em}
\setlength{\parindent}{0pt}

\uniformfont I will first use the video to identify the town shown in the clip. From the coastal scenery and visible landmarks, I will narrow the location down to \texttt{Tofino} before looking up the whale-protection rule.

\toolcall{\uniformfont Extract frames from the video and inspect visible location cues.}

\toolresponse{\uniformfont The clip shows a prominent surfing destination in British Columbia, consistent with \texttt{Tofino}.}
\end{sharedprefixbox}

\divergepoint{\uniformfont After identifying \texttt{Tofino}, the trajectories diverge on how they handle the Southern B.C.\ special protection measure: the failed branch applies the stricter \texttt{400 m} rule directly, while the successful branch checks whether that special corridor actually covers waters immediately offshore of Tofino.}

\noindent
\begin{minipage}[t]{0.34\linewidth}
\vspace{0pt}
\begin{failedtrajbox}
\uniformfont
\raggedright
\setlength{\parskip}{0.25em}
\setlength{\parindent}{0pt}

\uniformfont I have identified the town as \texttt{Tofino}. Next I will check official 2024/2025 killer-whale protection pages and use the applicable vessel-distance rule for this location.

\toolcall{\uniformfont \texttt{web\_search:} official killer-whale measures near Tofino, 2024/2025}

\toolresponse{\uniformfont The search returns official pages mentioning a \texttt{400 m} approach distance in a protected Southern B.C.\ corridor.}

\uniformfont I will use that stricter regional measure as the answer for waters offshore of \texttt{Tofino}.

\finalanswer{\uniformfont \texttt{Tofino, 400 meters}}
\end{failedtrajbox}
\end{minipage}
\hfill
\begin{minipage}[t]{0.64\linewidth}
\vspace{0pt}
\begin{successtrajbox}
\uniformfont
\raggedright
\setlength{\parskip}{0.25em}
\setlength{\parindent}{0pt}

\uniformfont I have identified the town as \texttt{Tofino}. Next I will check whether the stricter Southern B.C.\ killer-whale protection corridor actually applies to waters immediately offshore of this town, rather than assuming that the regional \texttt{400 m} rule governs here.

\toolcall{\uniformfont \texttt{web\_search:} official killer-whale measures near Tofino, 2024/2025}

\toolresponse{\uniformfont The search returns official pages mentioning a \texttt{400 m} approach distance in a protected Southern B.C.\ corridor.}

\uniformfont I will then check the general regulation for killer whales in Pacific waters so I can compare the special regional measure against the baseline legal rule.

\toolcall{\uniformfont \texttt{web\_search:} Marine Mammal Regulations killer whales Pacific waters 200 m}

\toolresponse{\uniformfont The official Marine Mammal Regulations give a baseline minimum approach distance of \texttt{200 m} for killer whales in Pacific waters.}

\uniformfont I will verify that the Southern B.C.\ special corridor does not directly govern the waters immediately offshore of \texttt{Tofino}, so I will use the general \texttt{200 m} rule instead.

\finalanswer{\uniformfont \texttt{Tofino, 200 metres}}
\end{successtrajbox}
\end{minipage}

\smallskip

\uniformfont\noindent\textbf{Key Difference.}
The failed trajectory directly applies the Southern B.C.\ special \texttt{400 m} measure to Tofino, whereas the successful trajectory explicitly checks whether that special corridor covers Tofino's offshore waters before falling back to the general \texttt{200 m} rule.

\end{tcolorbox}

\endgroup

\caption{A representative OmniGAIA case comparing GPT-5.4 \texttt{high} without Skills and with Human-in-the-loop Skills. The no-Skills run answers incorrectly by directly applying the Southern B.C.\ special \texttt{400 m} rule to Tofino, whereas the Human-in-the-loop-Skills run answers correctly using \texttt{200 m} rule for the waters immediately offshore of Tofino.}
\label{fig:idx161_tofino_case}
\end{figure}

%% file: Sections/5-training.tex
\section{\method: Towards Stronger Coding Agent for Omnimodal Content}
\label{sec:codex}

\begin{figure}[!h]
\centering

\begin{minipage}[t]{0.65\linewidth}
    \vspace{0pt}
    \centering
    \scriptsize
    \setlength{\tabcolsep}{3pt}
    \resizebox{\linewidth}{!}{%
    \begin{tabular}{lrrrl}
    \toprule[1.5pt]
    Subset & All & SFT & RL & Type / Modality \\
    \midrule
    \multicolumn{5}{l}{\emph{By source}} \\ 
    \midrule
    Omnimodal-Agent-SFT-2K & 2,110 & 1,413 & 697 & open-ended, multimodal \\
    OmniVideoBench & 997 & 670 & 327 & MCQ, video+audio \\
    AVUTBenchmark & 1,700 & 1,136 & 564 & MCQ, video+audio \\
    Video-MME-v2 selected & 1,228 & 823 & 405 & MCQ, video \\
    \midrule
    \multicolumn{5}{l}{\emph{By modality}} \\ 
    \midrule
    Video+audio & 2,369 & 1,585 & 784 & dual-stream media \\
    Video & 2,225 & 1,493 & 732 & visual-temporal media \\
    Audio+image & 1,437 & 961 & 476 & cross-modal media \\
    Audio & 4 & 3 & 1 & audio-only media \\
    \midrule
    \textbf{Total} & \textbf{6,035} & \textbf{4,042} & \textbf{1,993} & \\
    \bottomrule[1.5pt]
    \end{tabular}
    }
    \captionof{table}{Statistics of \textsc{OmniCoding}.}
    \label{tab:dataset_stats}
\end{minipage}
\hfill
\begin{minipage}[t]{0.33\linewidth}
    \vspace{0pt}
    \centering
    \includegraphics[width=\linewidth]{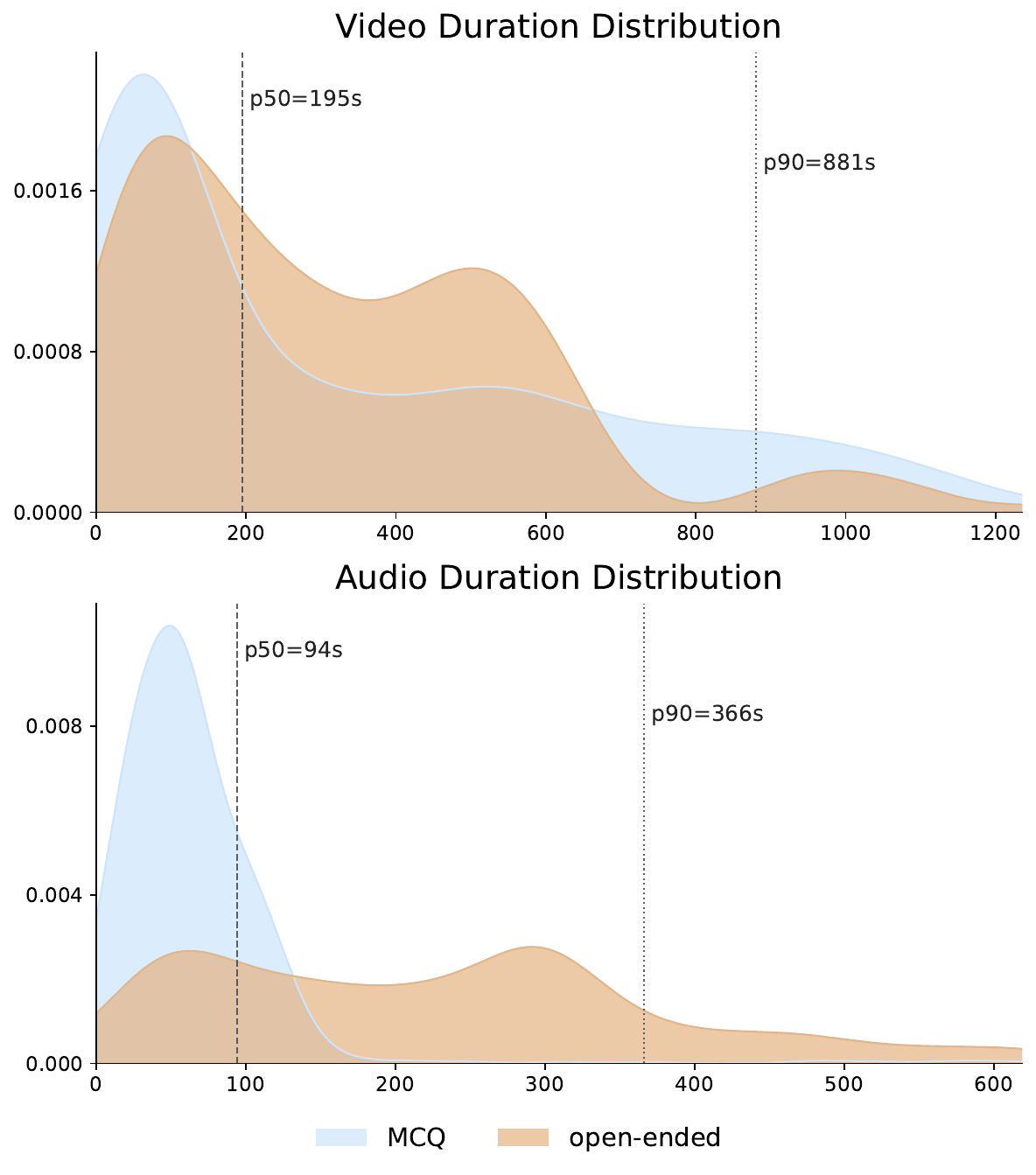}
    \captionof{figure}{Duration distribution.}
    \label{fig:duration_density}
\end{minipage}

\end{figure}

Given that closed-source coding models outperform native omni-modal models in both performance and efficiency, we ask whether such capability can be elicited through post-training and reproduced on open-source models. To this end, we propose a strong baseline on Qwen-3.5-9B and Qwen-3.6-27B, with a complete SFT-then-RL training recipe, aiming to explore this direction on open-source coding agents on omnimodal content.

\subsection{\dataset: Synthesized Omnimodal Coding Agent Trajectory Dataset}
We construct \textsc{OmniCoding}, a dataset of 6,035 verifiable question--answer examples paired with video, audio, or image files for multimodal terminal-tool-use agents. Each example follows a unified schema with a question, a list of acceptable ground-truth answers, media paths, and source/modality/task metadata. We draw from four complementary sources, OmniGAIA-SFT-2K \citep{li2026omnigaia}, OmniVideoBench \citep{li2025omnivideobench}, AVUTBenchmark \citep{yang2025avut}, and Video-MME-v2 \citep{fu2026video}, keeping only the Video-MME-v2 subset requiring audio-visual or temporal reasoning. Table~\ref{tab:dataset_stats} and Figure~\ref{fig:duration_density} report the per-source composition and media-duration distribution. Construction proceeds in three steps: \textbf{(i)~Normalization}, unifying all sources into the same schema with deduplicated questions and media consolidated into a flat directory; \textbf{(ii)~Verifiability}, regenerating open-ended questions with ground-truth lists covering common surface variants (units, abbreviations, capitalization, number formats) and converting multiple-choice into option-letter answer sets; and \textbf{(iii)~Cross-review}, verifying answer lists and task coverage. We stratified-split the data into 4,042 SFT and 1,993 RL examples by source-task key, and use exact match against the ground-truth list (extracted from an \texttt{<answer>} tag) as the binary RL reward.

\subsection{Method and Training Infrastructure}
\label{sec:rl-training}

We train the terminal agent with supervised warm-start followed by reinforcement learning from verifiable rewards. All training examples follow the schema $x=(q,\mathcal{F},\mathcal{Y})$, where multiple-choice tasks use option-letter answer sets and open-ended tasks are assigned curated answer sets covering common aliases, units, abbreviations, capitalization, and number formats. The agent submits its final answer inside \verb|<answer>|.

\header{Supervised warm start.}
For examples with demonstration trajectories, we fine-tune on assistant-generated tokens only, masking terminal outputs, tool responses, and environment observations from the loss. This stage teaches the basic interaction protocol: inspect files, run commands, create intermediate artifacts, recover from tool failures, and submit answers in the required format.

\header{Reward design.}
We design a process-aware verifiable reward that goes beyond binary answer matching, addressing three failure patterns: (1)~the agent guesses without engaging the workspace, (2)~it ignores the relevant modality (e.g., answering a video question from text alone), and (3)~it emits malformed, escape-attempting, or otherwise broken tool calls. For each rollout $\tau$ we compute
\begin{equation*}
R(x,\tau) =
\begin{cases}
0, & \text{if } \tau \text{ is removed (timeout / step-limit / context overflow)}, \\[2pt]
\text{base}(\tau) + r_{\mathrm{mod}}(\tau) + r_{\mathrm{tool}}(\tau), & \text{otherwise},
\end{cases}
\end{equation*}
with the following components:
\begin{itemize}[leftmargin=*,itemsep=1pt]
    \item \textbf{Three-tier base reward.} $\text{base}(\tau)=w_{\text{correct}}$ if gated-correct (normalized exact match against $\mathcal{Y}$, modality engaged, $\ge 1$ tool call, no escape attempt); else $w_{\text{fmt}+}$ if the agent terminated cleanly via \texttt{task\_complete} with a non-empty \verb|<answer>|; else $w_{\text{fmt}-}$. The format bonus does \emph{not} stack with correctness; it exists solely to keep gradient signal on tried-correctly-but-answered-wrong trajectories that would otherwise collapse the group to zero variance.
    \item \textbf{Modality penalty $r_{\mathrm{mod}}$.} A fixed negative weight $w_{\mathrm{mod}}$ is charged when the task contains video or audio but the agent invokes no tool from the corresponding universe (\texttt{ffmpeg}, \texttt{whisper}, etc.), detected by substring matching on bash commands.
    \item \textbf{Bad-tool penalty $r_{\mathrm{tool}}$.} A weighted ratio $-w_{\mathrm{tool}}\cdot P_{\text{bad}}$ with $P_{\text{bad}}=(n_{\text{unparseable}}+n_{\text{disallowed}}+n_{\text{escape}}+n_{\text{syntax-fail}})/(1+n_{\text{tool}})$ aggregates malformed calls, disallowed tools, gold-answer leakage attempts, and persistent shell errors.
\end{itemize}
Weight values are listed in Appendix~\ref{appendix:reward_hparams}. We further apply group-level mechanisms: \textbf{(i)~group advantage normalization} centers rewards within each group of $G$ sibling rollouts; \textbf{(ii)~zero-variance masking} drops groups where all rollouts succeed or fail; \textbf{(iii)~active-only baseline} excludes removed samples from group statistics; and \textbf{(iv)~dynamic filter} drops groups with $\le 1$ active sample or near-zero variance originated from DAPO \citep{yu2025dapo}.

\header{GSPO update.}
For each task we sample $G$ rollouts $\tau_i\sim\pi_{\theta_{\mathrm{old}}}(\cdot\mid x)$ and compute group-normalized advantages $A_i=(R_i-\bar{R})/(\sigma_R+\epsilon_A)$. Let $z_i=(z_{i,1},\ldots,z_{i,L_i})$ be the model-generated tokens in $\tau_i$ (tool outputs masked from the loss) and $c_{i,\ell}$ the transcript prefix before $z_{i,\ell}$. Following GSPO~\citep{zheng2025gspo}, we use a length-normalized sequence-level importance ratio and optimize the clipped objective:
\begin{equation*}
\resizebox{\linewidth}{!}{$
\displaystyle
\rho_i(\theta)=
\exp\!\left(\frac{1}{L_i}\sum_{\ell=1}^{L_i}
\big[\log\pi_\theta(z_{i,\ell}\mid c_{i,\ell})
-\log\pi_{\theta_{\mathrm{old}}}(z_{i,\ell}\mid c_{i,\ell})\big]\right),
\qquad
\mathcal{J}_{\mathrm{GSPO}}(\theta)=
\mathbb{E}_{x}\!\left[\frac{1}{G}\sum_{i=1}^{G}
\min\!\Big(\rho_i(\theta) A_i,\,
\mathrm{clip}\!\big(\rho_i(\theta),1-\epsilon_-,1+\epsilon_+\big)A_i\Big)\right].
$}
\end{equation*}
Since reward is assigned to whole trajectories, sequence-level clipping is a natural fit: one importance ratio weights all generated tokens while tool outputs stay masked.

\header{Training infrastructure.}
Our RL system decouples sandboxed rollout generation from policy optimization. Rollout workers run the current policy in isolated workspaces, execute terminal commands, collect trajectories, parse answers, compute reward components, and return old log-probabilities and masks; the trainer consumes these logs and applies GSPO without executing tools or touching raw media. This separation is useful because omni-modal rollouts involve non-differentiable operations, while the optimizer only needs token sequences, masks, old log-probabilities, and scalar rewards. Sandboxes are hosted on a Slurm cluster with CPU/GPU nodes and storage, the trainer runs on cloud servers, and the two are connected via a \emph{Cloudflare} tunnel for trajectory collection.

\subsection{Experiment Setups and Results}

\header{Agent Harness.}
To train \textsc{OmniCoding}, we build a local terminal-agent harness adapted from Terminus-KIRA~\citep{terminuskira2026} for converting raw multimodal files into small, checkable agent operations. At each episode, the environment creates an isolated workspace with the question and staged media files. The media remains in the filesystem rather than being packed into the model context, so the agent must decide how to inspect it through audio extraction, frame sampling, OCR/ASR, metadata reading, scripting, and intermediate verification. The visible action space is deliberately compact: \texttt{execute\_commands} runs one command in a persistent shell, \texttt{image\_read} inspects an image as a native multimodal observation, and \texttt{task\_complete} ends the episode. Higher-level media processing is expressed through code and command-line tools rather than benchmark-specific APIs, keeping the harness lightweight but expressive. Unlike general software-engineering harnesses~\citep{wang2024openhands,minisweagent2025}, our scaffold focuses on offline multimedia processing and trajectory collection: tool calls, terminal/image observations, final answers, exit reasons, and rewards are serialized for SFT filtering, RL replay, and failure analysis. Full details are in Appendix~\ref{app:omnicoding-harness}.

\header{Baselines and Evaluation Protocol.} We train Qwen3.5-9B and Qwen3.6-27B and compare against three baselines: \textbf{(i)}~training-free direct MLLM inference, where the model is fed visual frames sampled at fps=1 (capped at 512) together with Whisper-large ASR transcripts; \textbf{(ii)}~the same MLLM evaluated under our harness with direct image reading; and \textbf{(iii)}~SFT warm-up baseline. We additionally include Codex GPT-5.4 (low) under our harness as a reference. The thinking budget is set to 4K tokens; if the model fails to produce a correctly formatted answer, we issue a multi-turn reminder to re-answer in the required format, and if reasoning is truncated, the model is allowed to continue thinking. We evaluate on the same benchmark suite as before, reporting pass@1 throughout.

\header{Experiment Results and Analysis.}
As shown in Table \ref{tab:omnicoding_main}, model-tool-environment finetuned models are more adaptive to harness and perform much better than baseline and even direct MLLM settings that take up to 512 images. \method-RL-9B shows extremely reasoning and terminal tool-use capability, achieving 15.8 on VideoZeroBench, which is almost competitive to Gemini 3.1 Pro.

\begin{table*}[!t]
\centering
\caption{Performance of Qwen3.5-9B and Qwen3.6-27B under different settings on four benchmarks. For the Direct MLLM setting in the OmniGAIA benchmark, we provide models with tools following their codebase. The thinking budget is fixed at 4K tokens for all settings.}
\label{tab:omnicoding_main}
\vspace{-0.5em}
\small
\scalebox{0.7}{%
\begin{tabular}{l | cccc | cc | cccc | c}
\toprule[1.5pt]
\textbf{\multirow{2}{*}{Model/Setting}}
& \multicolumn{4}{c|}{\textbf{OmniGAIA}}
& \multicolumn{2}{c|}{\textbf{Social Omni}}
& \multicolumn{4}{c|}{\textbf{LVOmniBench}}
& \textbf{VideoZeroBench} \\
\cmidrule(lr){2-5} \cmidrule(lr){6-7} \cmidrule(lr){8-11} \cmidrule(lr){12-12}
& Easy & Medium & Hard & Avg.
& Level1 & Level2
& Easy & Medium & High & Avg.
& Level 3 \\
\midrule
\multicolumn{12}{c}{\textit{\textbf{Strong Baselines}}} \\
\midrule
\rowcolor{gray!20}  OmniAtlas-Omni-30B-A3B & 31.1 & 18.8 & 9.0 & 20.8 & 50.0 & 61.0 & 44.4 & 23.1 & 36.0 & 34.0 & 6.0 \\
\rowcolor{gray!20}  Codex-GPT-5.4 Low & 63.1 & 53.8 & 44.9 & 55.0 & 64.0 & 59.0 & 58.3 & 46.2 & 36.0 & 48.0 & 21.0 \\
\midrule
\multicolumn{12}{c}{\textit{\textbf{Qwen3.5-9B}}} \\
\midrule
Direct MLLM       & 23.0 & 16.2 & 23.1 & 20.0 & 37.0 & 59.0 & 44.4 & 33.3 & 16.0 & 33.0 & 7.2 \\
Baseline (direct image)        & 23.8 & 14.4 & 17.9 & 18.3 & 41.0 & 58.0 & 41.7 & 15.4 & 36.0 & 30.0 & 5.9 \\
\method (Ours, SFT) & 30.0 & 21.0 & 17.0 & 23.1 & 50.0 & 61.0 & 48.6 & 28.9 & 16.0 & 33.0 & 9.5 \\
\method (Ours, RL) & 32.8 & 26.3 & 24.4 & 28.1 & 52.0 & 63.0 & 72.2 & 41.0 & 44.0 & 53.0 & 15.8 \\
\textcolor{deeppink}{\textit{$\Delta$ vs Direct MLLM}} & \textcolor{deeppink}{\textit{+42.6\%}} & \textcolor{deeppink}{\textit{+62.3\%}} & \textcolor{deeppink}{\textit{+5.6\%}} & \textcolor{deeppink}{\textit{+40.5\%}} & \textcolor{deeppink}{\textit{+40.5\%}} & \textcolor{deeppink}{\textit{+6.8\%}} & \textcolor{deeppink}{\textit{+62.6\%}} & \textcolor{deeppink}{\textit{+23.1\%}} & \textcolor{deeppink}{\textit{+175.0\%}} & \textcolor{deeppink}{\textit{+60.6\%}} & \textcolor{deeppink}{\textit{+119.4\%}} \\
\midrule
\multicolumn{12}{c}{\textit{\textbf{Qwen3.6-27B}}} \\
\midrule
Direct MLLM & 44.3 & 38.1 & \textbf{28.6} & 38.2 & 61.0 & 47.0 & 66.7 & 46.2 & 36.0 & 51.0   & 7.4 \\
Baseline (direct image) & 32.8 & 18.8 & 23.1 & 24.4 & 46.0 & 55.0 & 52.8 & 33.3 & 32.0 & 40.0 &  13.0 \\
\method (Ours, SFT) & \textbf{50.0} & \textbf{45.0} & 28.2 & \textbf{43.3} & \textbf{64.0} & \textbf{67.0} & \textbf{75.0} & \textbf{53.8} & \textbf{48.0} & \textbf{60.0} & \textbf{18.4} \\
\textcolor{deeppink}{\textit{$\Delta$ vs Direct MLLM}} & \textcolor{deeppink}{\textit{+12.9\%}} & \textcolor{deeppink}{\textit{+18.1\%}} & \textcolor{deeppink}{\textit{$-$1.4\%}} & \textcolor{deeppink}{\textit{+13.4\%}} & \textcolor{deeppink}{\textit{+4.9\%}} & \textcolor{deeppink}{\textit{+42.6\%}} & \textcolor{deeppink}{\textit{+12.4\%}} & \textcolor{deeppink}{\textit{+16.5\%}} & \textcolor{deeppink}{\textit{+33.3\%}} & \textcolor{deeppink}{\textit{+17.6\%}} & \textcolor{deeppink}{\textit{+148.6\%}} \\
\bottomrule[1.5pt]
\end{tabular}
}
\vspace{-1em}
\end{table*}

%% file: Sections/6-benchmark.tex
\vspace{-0.5em}
\section{\benchmark: From Many-modality Understanding to Processing}
\label{sec:terminalbencho}
\vspace{-0.5em}
\begin{figure*}[!t]                                            
    \centering     
    \includegraphics[width=\linewidth]{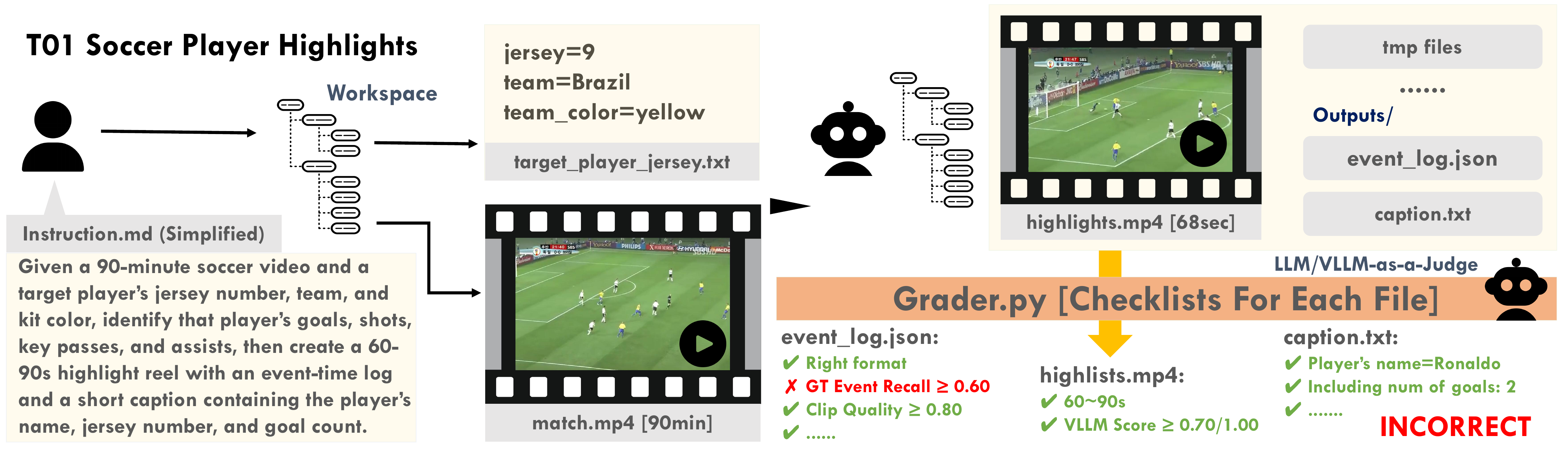}
    \vspace{-1.5em}
    \caption{Overview of our benchmark, illustrated with Task T01. The coding agent is required to generate a highlight clip and caption from a soccer video and a player query, and its outputs are evaluated by an LLM-based judge in terms of event accuracy, video quality, and task correctness.}  
    \label{fig:benchmark_case_study}  
    \vspace{-1em}                                      
\end{figure*}

As many-modality understanding capabilities mature, the frontier is shifting from passive \textit{understanding} to active \textit{processing}: editing videos, auditing image databases, and producing spec-compliant multimodal outputs. This shift reveals a gap between native omni-models, which excel at end-to-end perception but cannot directly operate on files, and coding agents, which can decompose tasks into tool-use pipelines. Yet existing benchmarks do not evaluate this capability: omnimodal benchmarks such as OmniGAIA~\citep{li2026omnigaia} and LVOmniBench~\citep{tao2026lvomnibench} focus on QA-based perception, while coding-agent benchmarks such as Terminal-Bench~\citep{merrill2026terminal} and Claw-Eval~\citep{ye2026claw} mainly target text or text+image tasks. As shown in Table~\ref{tab:benchmark_comparison}, none jointly cover many-modality understanding (MMU) and many-modality processing (MMP) on real-world cases. To close this gap, we introduce \benchmark, the first process-level benchmark for coding agents on omnimodal tasks, requiring agents to process realistic inputs such as long videos, multilingual documents, and noisy audio through multi-stage pipelines, and evaluating both multimodal understanding and operation capability.

\header{Design Principles.}
\benchmark comprises 50 challenging tasks across video (38\%), audio (26\%), image (14\%), document (4\%), and cross-modal (18\%) categories. Four core principles guide the design: \textbf{(i)~Real-world cases}, where all tasks are drawn from authentic sources (Internet Archive, HuggingFace datasets, public videos) rather than synthetic constructions; \textbf{(ii)~Multi-dimensional grading}, where each task is scored along 4--8 weighted dimensions covering accuracy, cross-document consistency, and output quality; \textbf{(iii)~Shortcut-resistant design}, which combines closed vocabularies, cross-file ID consistency checks, decoy injection, and word limits to prevent gaming; and \textbf{(iv)~Process-level evaluation}, which validates intermediate artifacts and reasoning chains rather than only final answers.

\header{Task Categories.}
Representative tasks include: \textit{Video Understanding \& Processing} (T01a: extract 60-90s highlights from 90-min soccer broadcast by jersey number; T03a: auto-edit 5-min final cut from 3-camera meeting footage); \textit{Audio Forensics} (T107a: segment 22-min concert recording into songs with tempo/defect detection; T108a: extract rhetorical devices and thematic structure from political speech); \textit{Image Database Analysis} (T13a: detect wood defects with pixel-level masks; T50a: audit mislabeled images in CIFAR-10 subset). Unlike QA benchmarks, \benchmark requires agents to \textit{create} multimodal outputs (edited videos, text+video interleaved reports) that satisfy strict format and consistency constraints.

\begin{table}[!t]
\centering
\caption{Overall accuracy and average score of different coding agents on \benchmark. \textsuperscript{$\dagger$}Models without native vision capabilities use the image-read implemented with Qwen3.5-Flash.}
\label{tab:clawbench_model_eval_summary}
\vspace{-0.5em}
\scalebox{0.9}{%
\begin{tabular}{l | c | cc}
\toprule[1.5pt]
\textbf{Model} & \textbf{Harness} & \textbf{Pass Rate (\%)} & \textbf{Avg. Score} \\
\midrule
GPT-5.5 Low                      & Codex & \textbf{24} & 0.68 \\
GPT-5.5 Medium                   & Codex & 18 & 0.69 \\
GPT-5.5 High                     & Codex & \textbf{24} & \textbf{0.71} \\
GPT-5.5 xHigh                    & Codex & \textbf{24} & \textbf{0.71} \\
GLM-5.1 (V)\textsuperscript{$\dagger$}               & CC & 16 & 0.69 \\
DeepSeek V4 Pro (V)\textsuperscript{$\dagger$}       & CC & 12 & 0.64 \\
DeepSeek V4 Flash (V)\textsuperscript{$\dagger$}     & CC & 10 & 0.60 \\
Kimi K2.6                                             & CC &  8 & 0.56 \\
MiniMax M2.7 (V)\textsuperscript{$\dagger$}          & CC &  4 & 0.55 \\
\bottomrule[1.5pt]
\end{tabular}
}
\end{table}

\header{Evaluation Protocol.}
Each task defines a weighted scoring formula $\text{score} = \sum_i w_i \cdot m_i$ where $m_i$ are normalized metrics (IoU, F1, accuracy, LLM-judged quality). A submission passes if $\text{score} \geq 70$ \textit{and} all gating conditions hold: required files exist, JSON schemas validate, cross-document IDs match, and outputs comply with closed vocabularies. Gating failures result in zero score regardless of partial correctness. For open-ended outputs (summaries, captions), we use LLM judges with detailed rubrics; for temporal tasks (video segmentation, audio boundaries), we apply tolerance windows (±2s) and Hungarian matching for IoU computation.

\header{Results Analysis.}
As shown in Table~\ref{tab:clawbench_model_eval_summary}, GPT-5.5 obtains the strongest results among the evaluated settings, with high and xHigh achieving the highest average scores. MiniMax M2.7 (V) performs the worst, reaching only a 4\% pass rate and an average score of 0.55. Overall, the results show that \benchmark remains challenging across different model and harness configurations, especially for tasks requiring reliable long-horizon tool use.

\begin{table}[t]
\centering
\caption{Comparison of \benchmark with existing benchmarks.}
\label{tab:benchmark_comparison}
\vspace{0em}
\small
\resizebox{\columnwidth}{!}{%
\begin{tabular}{l l l l c c c c}
\toprule[1.5pt]
\textbf{Benchmark} & \textbf{Modalities} & \textbf{\# Size} & \textbf{Target Capability} & \textbf{Real} & \textbf{Cross-source} & \textbf{MMU} & \textbf{MMP} \\
\midrule
OmniGAIA~\citep{li2026omnigaia} & V+A+I+T & 360 & QA, factuality seeking & \halfmark & \xmark & \cmark & \xmark \\
Terminal-Bench 2.0~\citep{merrill2026terminal} & T+F & 89 & Terminal workflows & \cmark & \halfmark & \xmark & \halfmark \\
Claw-Eval~\citep{ye2026claw} & T+I+V+D & 300 & Agent workflows & \cmark & \halfmark & \halfmark & \halfmark \\
WildClawBench~\citep{wildclawbench} & T+I+V+A+D & 60 & In-the-wild agent tasks & \cmark & \halfmark & \halfmark & \halfmark \\
\midrule
\textbf{\benchmark (Ours)} & \textbf{T+V+A+I+D+F} & \textbf{50} & \textbf{Many-modality Processing} & \textbf{\cmark} & \textbf{\cmark} & \textbf{\cmark} & \textbf{\cmark} \\
\bottomrule[1.5pt]
\end{tabular}
}
\vspace{0.2em}
\begin{minipage}{0.98\columnwidth}
\footnotesize
\cmark~denotes full support, \halfmark~denotes partial or mixed support, and \xmark~denotes absent support. MMU and MMP mean many-modality understanding and processing.
\textbf{V}: video, \textbf{A}: audio, \textbf{I}: image, \textbf{T}: text, \textbf{D}: document, \textbf{F}: files.
\end{minipage}
\vspace{-1.5em}
\end{table}

%% file: Sections/2-related.tex
\vspace{-0.5em}
\section{Related Works}
\vspace{-0.5em}
\header{Omnimodal Models and Evaluation}
Recent multimodal foundation models have advanced from single-modality extensions to omni-modal systems that jointly process speech, audio, images, and video \citep{chu2024qwen2audio, sun2024videosalmonn, xu2025qwen25omni, qwen35Omni, google2026gemini31pro}. As models push toward unified vision-and-speech interaction, evaluation has correspondingly shifted from task-specific video or audio benchmarks \citep{li2023mvbench, liu2024tempcompass, fu2024videomme, wu2024longvideobench, li2022avqa} to omni-modal suites that more explicitly target joint reasoning over visual, auditory, and textual cues, with growing attention to long videos, grounded interaction, and real-world complexity \citep{li2024omnibench, hong2025worldsense, zhang2025omnieval, li2026omnigaia, radwan2026sonico1, tao2026lvomnibench, xie2026socialomni}. Yet our audit shows that many supposedly audio-visual tasks still allow strong shortcuts or rely only weakly on audio \citep{kim2025needtolisten}, motivating more careful benchmark construction.

\header{Multimodal Agentic Tool-Use and Sandboxed Agents.}
Tool use has become a central mechanism for extending LLM agents beyond their parametric knowledge \citep{yao2022react, schick2023toolformer, qin2023toolllm, patil2025bfcl, yao2024taubench, wang2025mcpbench}, and increasingly improves multimodal understanding via active search, code generation, and tool orchestration without retraining a monolithic model \citep{hu2024visualsketchpad, zhao2025pyvision, zhang2025thyme, zhang2024omagent, zhang2025deepvideodiscovery, tao2025omniagent, lin2025agentomni, li2026lenswalk}. The most mature form of sandboxed agency has emerged in software engineering, where models resolve real issues inside an execution environment \citep{jimenez2023swebench, yang2024sweagent, wang2024openhands, kwa2025measuringlongsoftware, cheng2026llminsandbox}, but existing work primarily targets natively textual or computational tasks. Our work takes a step beyond: we show, for the first time, that a model can use a sandboxed terminal with self-constructed tool use to solve non-native multimodal tasks, outperforming many specialized omni models and suggesting that sandboxed agency is a general mechanism for broader many-modality problem-solving.

%% file: Sections/10-conclusion.tex
\section{Conclusion}
We study offline omnimodal understanding through the lens of sandboxed coding agents in both training-free and fine-tune settings. Across multiple benchmarks, these agents solve a surprisingly large fraction of video-audio tasks by selectively retrieving and processing evidence from raw inputs, often rivaling or surpassing native omnimodal systems. We hope this work motivates a deeper study of multimodal intelligence at the model–tool–environment interface.

\section*{Ackownledgement}
We thank Chenlong Wang, Jieneng Chen, Xirui Li, Ming Li, Kwesi Cobbina and Misaki Wang for their invaluable support and discussion. Dongping Chen is supported by Modal, Tavily, and Linear's academic research project.

%% file: Sections/99-appendix.tex
\clearpage
\input{Sections/3-definition}

\input{Sections/7-discussion}

\clearpage

\section{Full Related Works}

\subsection{Multimodal Models and Evaluation}
Recent progress toward omni-modal assistants reflects a shift from modality-specific perception modules to unified models that can process speech, general audio, images, and video within a single system. On the model side, audio-language systems such as Qwen2-Audio \citep{chu2024qwen2audio} and video-centered models such as VideoLLaMA~2 \citep{cheng2024videollama2} and video-SALMONN \citep{sun2024videosalmonn} established strong foundations for audio-visual understanding. More recent omni-modal systems, including VITA \citep{fu2024vita}, OpenOmni \citep{luo2025openomni}, Baichuan-Omni \citep{li2024baichuanomni}, Moshi \citep{defossez2024moshi}, and Qwen2.5-Omni \citep{xu2025qwen25omni}, move toward unified and often real-time interaction across speech and vision. These systems substantially improve the practicality of omni interaction, but they also make it harder to separate intrinsic model ability from gains due to decoding strategies, longer context, or external processing pipelines.

Benchmarking has evolved in parallel. Earlier suites mainly targeted either video understanding, such as MVBench \citep{li2023mvbench}, Video-MME \citep{fu2024videomme}, and MLVU \citep{zhou2024mlvu}, or audio / audio-visual understanding, such as AIR-Bench \citep{yang2024airbench}, MMAU \citep{sakshi2024mmau}, and AV-Odyssey \citep{gong2024avodyssey}. While these benchmarks advanced multimodal evaluation, many still under-represent audio, weakly couple audio with video, or emphasize short-context perception more than long-horizon multimodal reasoning.

Recent work therefore moves toward explicitly omni-modal and long-form evaluation. Benchmarks such as OmniBench \citep{li2024omnibench}, OmniEval \citep{zhang2025omnieval}, OmniGAIA \citep{li2026omnigaia}, LVOmniBench \citep{tao2026lvomnibench}, SocialOmni \citep{xie2026socialomni}, MMOU \citep{goel2026mmou}, and OmniPlay \citep{bie2025omniplay} target more realistic settings that require integrating visual, auditory, and textual evidence over longer horizons. At the same time, recent audits show that several audio-video benchmarks admit strong visual shortcuts \citep{kim2025needtolisten}, suggesting that progress in omni-modal evaluation depends not only on stronger models but also on shortcut-resistant benchmark design and explicit measurement of true modality dependence.

\subsection{Agentic Tool-Use}
The literature on tool-augmented language models treats external tools as a way to expand the action space of models rather than as simple add-ons. Foundational work such as ReAct \citep{yao2022react} and Toolformer \citep{schick2023toolformer} showed that reasoning can be interleaved with actions and that models can learn when to invoke tools. Later efforts, including Gorilla \citep{patil2023gorilla}, ToolBench \citep{xu2023toolbench}, ToolLLM \citep{qin2023toolllm}, BFCL \citep{patil2025bfcl}, and $\tau$-bench \citep{yao2024taubench}, scaled this paradigm to large tool collections and more interactive, stateful environments. Across these works, the central shift is from static capability elicitation to environment-mediated problem solving.

Executable code has emerged as one of the most expressive forms of tool use. CodeAct \citep{wang2024codeact} argues that executable programs provide a general interface for acting in digital environments, allowing agents to manipulate files, call libraries, and compose multi-step behaviors more flexibly than fixed action vocabularies. In multimodal settings, Visual Sketchpad \citep{hu2024visualsketchpad} and MLLM-Tool \citep{wang2024mllmtool} show that models can externalize perception into sketches, programs, and structured tool calls. PyVision \citep{zhao2025pyvision} and Thyme \citep{zhang2025thyme} further demonstrate that models can synthesize and execute Python-based image operations on demand, blurring the boundary between model capability and environment capability.

This perspective is increasingly visible in video and omni-modal reasoning. OmAgent \citep{zhang2024omagent}, OmniAgent \citep{tao2025omniagent}, Agent-Omni \citep{lin2025agentomni}, and LensWalk \citep{li2026lenswalk} all treat multimodal understanding as an active process of decomposition, search, and selective inspection rather than passive end-to-end perception. LongShOTBench \citep{kurpath2025longshotbench} explicitly evaluates tool use and iterative refinement in long-video settings. More broadly, LLM-in-Sandbox argues that rich environments can elicit capabilities not apparent in static chat settings, reinforcing the view that multimodal competence may emerge from the interaction between a strong base model and an expressive tool substrate \citep{cheng2026llminsandbox}. Together, these works motivate viewing multimodal intelligence as a property of a model--environment system rather than of a standalone backbone.

\subsection{Terminal Coding Agents and Benchmarks}

The most mature form of sandboxed agency has so far emerged in software engineering. SWE-bench \citep{jimenez2023swebench} established the canonical issue-resolution setting by grounding evaluation in real GitHub issues and executable repository states. Systems such as SWE-agent \citep{yang2024sweagent}, AutoCodeRover \citep{zhang2024autocoderover}, Agentless \citep{xia2024agentless}, and OpenHands \citep{wang2024openhands} explored different agent designs for this setting, ranging from search-heavy repair to lightweight execution-driven pipelines. This line of work reframes coding as long-horizon environment interaction rather than static code generation.

Benchmarks have expanded accordingly. SWE-bench Verified \citep{openai2024swebenchverified} improves reliability through a filtered subset, while SWE-bench Multimodal \citep{yang2024swebenchmultimodal} extends evaluation to visually grounded software issues. Terminal-Bench \citep{merrill2026terminalbench} and LongCLI-Bench \citep{feng2026longclibench} move beyond repository patching toward broader command-line tasks and longer-horizon CLI interaction. Other recent benchmarks, including SetupBench \citep{arora2025setupbench}, SWE-Bench Pro \citep{deng2025swebenchpro}, and SEC-bench \citep{lee2025secbench}, further broaden evaluation to environment setup, longer-horizon issue resolution, and security workflows.

The ecosystem has also expanded beyond benchmark construction into self-improvement and broader autonomy evaluation. SWE-smith \citep{yang2025swesmith} studies scalable data generation for software engineering agents, Live-SWE-agent \citep{xia2025livesweagent} explores self-evolution during deployment, and SWE-Fuse \citep{wen2026swefuse} uses trajectory learning to improve agent performance. Beyond software-specific tasks, RE-Bench \citep{wijk2024rebench}, HCAST \citep{rein2025hcast}, and TheAgentCompany \citep{xu2024theagentcompany} evaluate agents on longer-horizon work in realistic environments. At the same time, recent meta-analyses caution that benchmark realism, contamination, and task validity remain open concerns, and that high scores do not automatically imply broader engineering competence \citep{martinez2026whatsinbenchmark,prathifkumar2025swebenchmemory}. These critiques are especially relevant when interpreting terminal-agent performance as evidence of general agentic ability.

\section{\benchmark Construction Details}

TerminalBench-0 contains
50 curated terminal tasks under a shared workspace interface, each requiring the
agent to create concrete artifacts under \texttt{/workspace/output} rather than
return a short answer. 

\begin{table}[t]
\centering
\small
\setlength{\tabcolsep}{6pt}
\caption{Summary statistics for \benchmark.}
\label{tab:terminalbench0_stats}
\begin{tabular}{lr}
\toprule
Statistic & Value \\
\midrule
Tasks & 50 \\
Agent-visible fixture files & 3,013 \\
audio/video/image/PDF/FITS files & 1,293 \\
\bottomrule
\end{tabular}
\end{table}

\paragraph{Raw data.}
We ground the suite in real-source data: public research
datasets, public archives, public-domain scans, and direct-download media. The
sources cover broadcast/web video, AMI and legal speech, action and egocentric
video, industrial and scientific imagery, geospatial data, music stems,
bioacoustics, score-audio pairs, FITS files, and historical comics. Synthetic
content is reserved for controlled corruptions, decoys, overlays, and auxiliary
labels needed to make evaluation precise.

\paragraph{Preprocessing.}
Each source is normalized into a stable workspace representation before task
packaging. We also prepare
closed vocabularies, masks, catalogs, WCS headers, timestamp maps, and
source-to-fixture mappings for downstream grading.

\paragraph{Construction.}
\textbf{Step 1: Candidate mining.}
We first select source segments whose native structure supports a realistic
operation task, such as eventful video intervals, speaker turns, action clips,
score-aligned music excerpts. Candidate
segments are filtered for duration, visibility, signal quality, annotation
availability, and non-triviality before fixture generation.

\textbf{Step 2: Task-specific transformation.}
We then impose an operation or creation objective on the selected source. Some
tasks add controlled corruptions, hidden mappings, decoys, overlays, splice
events, synchronization errors, or multi-source audio compositions. Others keep
the media unchanged but require structured deliverables such as ledgers,
catalogs, action datasets, OMR alignments, reading-order tables, or redaction
reports.

\textbf{Step 3: Hidden reference synthesis.}
For each instance, we write hidden references that specify the answer geometry
used by the grader. These references include event windows, speaker maps,
transcripts, quote spans, crop paths, clean media, masks, stem targets,
timestamp curves, object tracks and
closed-vocabulary labels. They are withheld from the agent but used to score
both final outputs and intermediate artifacts.

\textbf{Step 4: Output-surface packaging.}
Finally, we expose each task as a small working directory rather than a
question-only prompt. The visible files may include media, manifests, closed
vocabularies, policy snippets, PDFs, CSVs, point clouds, FITS files, or
auxiliary metadata, and the expected outputs are ordinary files such as
MP4/WAV/PNG artifacts, CSV/JSON ledgers, subtitles,plots, and
reports. This format keeps the benchmark native to terminal tools,
\texttt{ffmpeg}, OCR/ASR, numerical Python, and file-system coordination.

\paragraph{Filtering.}
We apply filtering before and after packaging. Source candidates are removed if
they are too short, ambiguous, silent, weakly annotated, or trivially solvable. Packaged tasks are then checked for
complete schemas, aligned hidden references, active closed vocabularies, decoy
coverage, and resistance to shortcuts such as pass-through media, empty
manifests, constant labels, or copied templates. 

\paragraph{Evaluation.}
Each task uses a dedicated grader with weighted dimensions and gating
conditions. Deterministic checks cover temporal IoU, event recall,
closed-vocabulary accuracy, WER/CER, speaker attribution, ASR preservation,
signal metrics, SSIM/PSNR, mask IoU, timestamp error, catalog matching,
photometric error, geospatial overlap, table consistency, and schema validity.
For naturally open-form artifacts such as memos, overlays, or semantic
summaries, some graders add constrained LLM/MLLM checklist axes. The final score
therefore measures process-level completion: media quality, structured
intermediate artifacts, format compliance, and cross-document consistency.

\section{Failure Mode Taxonomy and Process-level Trajectory Dataset}
\label{appendix:failure_modes}
\begin{figure*}[!b]                                            
    \centering
    \includegraphics[width=\linewidth]{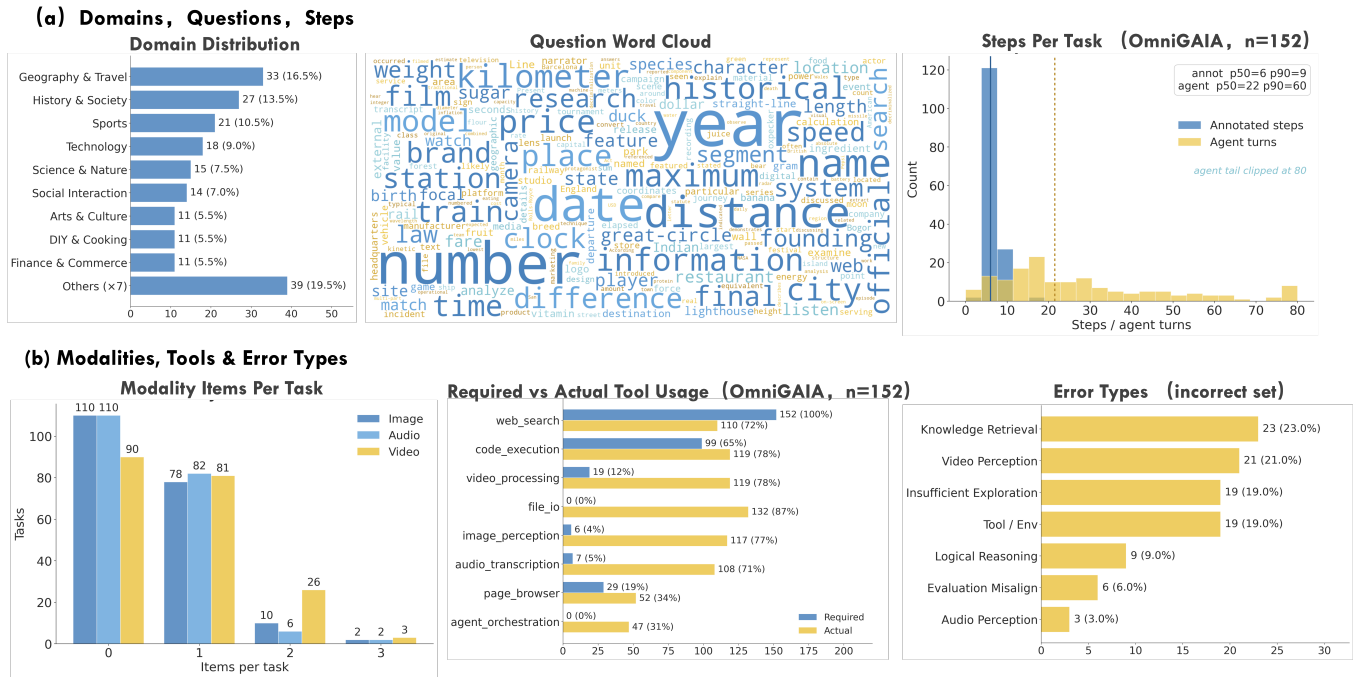}%
    \caption{%
      Overview of our process-level benchmark.                 
      \textbf{(a) Dataset Overview:} domain distribution across the union of OmniGAIA, LVOmniBench and SocialOmniBench (200 tasks); a word cloud of the 
  question text; the distribution of annotated logical steps per task overlaid with the agent's actual turn count.
      \textbf{(b) Capability Analysis:} per-task counts of image / audio / video inputs; required-versus-actual tool-category coverage on the OmniGAIA  
  subset (n=152) -- ``Required'' is drawn from \texttt{required\_external\_tools} $\cup$ \texttt{sources}, \textit{``Actual''} is recovered by argument-level    
  classification of the agent trajectory; and the distribution of primary error types on the 100 incorrect trajectories we annotate.
    }                                                                 \label{fig:bench_overview}                                       
\end{figure*}

Here we provide detailed definitions of the six failure modes introduced in the main text.

\begin{itemize}[leftmargin=*,itemsep=2pt]
    \item \textbf{Audio Perception and Extraction Error.} The agent mis-transcribes, mis-segments, or mis-attributes the audio signal itself (\emph{e.g.}, ASR errors under noise or speaker confusion), so all downstream reasoning is conditioned on content that diverges from the true recording.
    \item \textbf{Video Perception and Extraction Error.} The visual stream is misread at the source via blank or off-key frame sampling, OCR mistakes, or inverted event ordering, leaving later stages with incorrect evidence.
    \item \textbf{Insufficient Exploration of Modal Content.} With multimodal inputs available, the agent answers from textual cues or priors, or terminates exploration before critical evidence is verified. The failure lies in the decision to stop probing rather than in perception quality.
    \item \textbf{Knowledge Retrieval and Factual Error.} Given correct perception, the agent links it to wrong external knowledge through poorly specified queries, unreliable sources, conflated entities, or fabricated facts when retrieval fails.
    \item \textbf{Logical Reasoning and Calculation Error.} Evidence is largely correct, but the derivation breaks down via wrong formulas (\emph{e.g.}, planar vs.\ spherical distance), arithmetic mistakes, or dropped conditions in multi-step deductions.
    \item \textbf{Tool and Environment Infrastructure Failure.} The agent picks the right tool but is blocked by missing dependencies, sandbox permissions, or resource limits, forcing later steps to build on corrupted intermediates.
\end{itemize}

\begin{wraptable}{r}{0.5\textwidth}
    \centering
    \small
    \setlength{\tabcolsep}{6pt}
    \vspace{-1em}
    \caption{LLMs as process-level judges on our 200-trajectory subset.}
    \label{tab:reward_model_judge}
    \vspace{-0.5em}
    \begin{tabular}{lccc}
        \toprule
        Judge model & Step & Primary & Budget\\
        \midrule
        GPT-5.4-mini             & 0.588 & 0.657 & 3.17$\times$ \\
GPT-5.4                  & 0.737 & 0.830 & 10.54$\times$ \\
GPT-5.5                  & \textbf{0.746} & \textbf{0.840} & 32.49$\times$ \\
Qwen3.6-35B-A3B          & 0.605 & 0.813 & 1.68$\times$ \\
DeepSeek-V4-Flash        & 0.636 & 0.786 & 0.46$\times$ \\
        \bottomrule
    \end{tabular}
    \vspace{-1em}
\end{wraptable}

\label{appendix:llm_judge}

To scale failure mode analysis beyond the 200 human-annotated trajectories, we benchmark off-the-shelf LLMs as process-level judges and validate the chosen judge against human labels on this annotated set. Each model replays all 200 trajectories under a unified prompt, predicting a step reward $\in\{-1,0,+1\}$ for every assistant step and, on the 100 incorrect trajectories, a primary error type from our taxonomy. We follow the AgentProcessBench \citep{fan2026agentprocessbench} definition for process reward, with the rubric summarized below.

\begin{table}[h]
\centering
\small
\setlength{\tabcolsep}{4pt}
\caption{Step-type-specific criteria for the $\{-1,0,+1\}$ process reward.}
\label{tab:process_reward_step_types}
\renewcommand{\arraystretch}{1.25}
\begin{tabular}{p{2.6cm} p{3.4cm} p{3.4cm} p{3.4cm}}
\toprule
\textbf{Step type} & \textbf{$+1$ (effective)} & \textbf{$0$ (neutral)} & \textbf{$-1$ (harmful)} \\
\midrule
Text-only step (planning, reasoning) & Useful plan or constraint with concrete task value. & Generic narration or low-value planning. & States false facts or commits to a bad premise. \\
\midrule
Tool call & Tool choice is appropriate and materially useful. & Reasonable but low-yield, redundant, externally blocked, or based on incomplete information. & Clearly wrong tool, violates instructions, or repeats a failed pattern without strategy change. \\
\midrule
Post-tool interpretation & Interpretation is supported by the tool output and advances the task. & Cautious but inconclusive. & Misreads the output or overclaims beyond the evidence. \\
\midrule
Final answer / decisive claim & Supported by evidence already collected in the trajectory. & Cautious partial synthesis that does not materially help. & Guesses, hallucinates, or depends on an unresolved earlier error. \\
\bottomrule
\end{tabular}
\end{table}

\header{Step reward rubric.}
Only \textit{assistant} messages are scored; \textit{user} and \textit{tool} messages serve as evidence. Each assistant step is judged using only information visible up to that step (no hindsight), and the three-level label is defined as follows:
\begin{itemize}[itemsep=1pt, leftmargin=*]
    \item \textbf{$+1$ (correct and effective).} The step is factually correct given the available evidence and clearly advances the task, e.g., appropriate tool use, evidence-supported interpretation of a tool output, an effective decision, or an explicit correction of a prior mistake.
    \item \textbf{$0$ (neutral or exploratory).} The step is reasonable but yields limited or unclear progress, e.g., sensible exploration, tentative planning, redundant recap, or a reasonable tool attempt that fails for external reasons such as a 404, timeout, or tool limitation.
    \item \textbf{$-1$ (incorrect or harmful).} The step is factually wrong, misreads or fabricates evidence, violates system or policy constraints, repeats a failed pattern without a meaningful strategy change, or otherwise moves the trajectory away from success.
\end{itemize}

\begin{figure}
    \centering
    \includegraphics[width=1\linewidth]{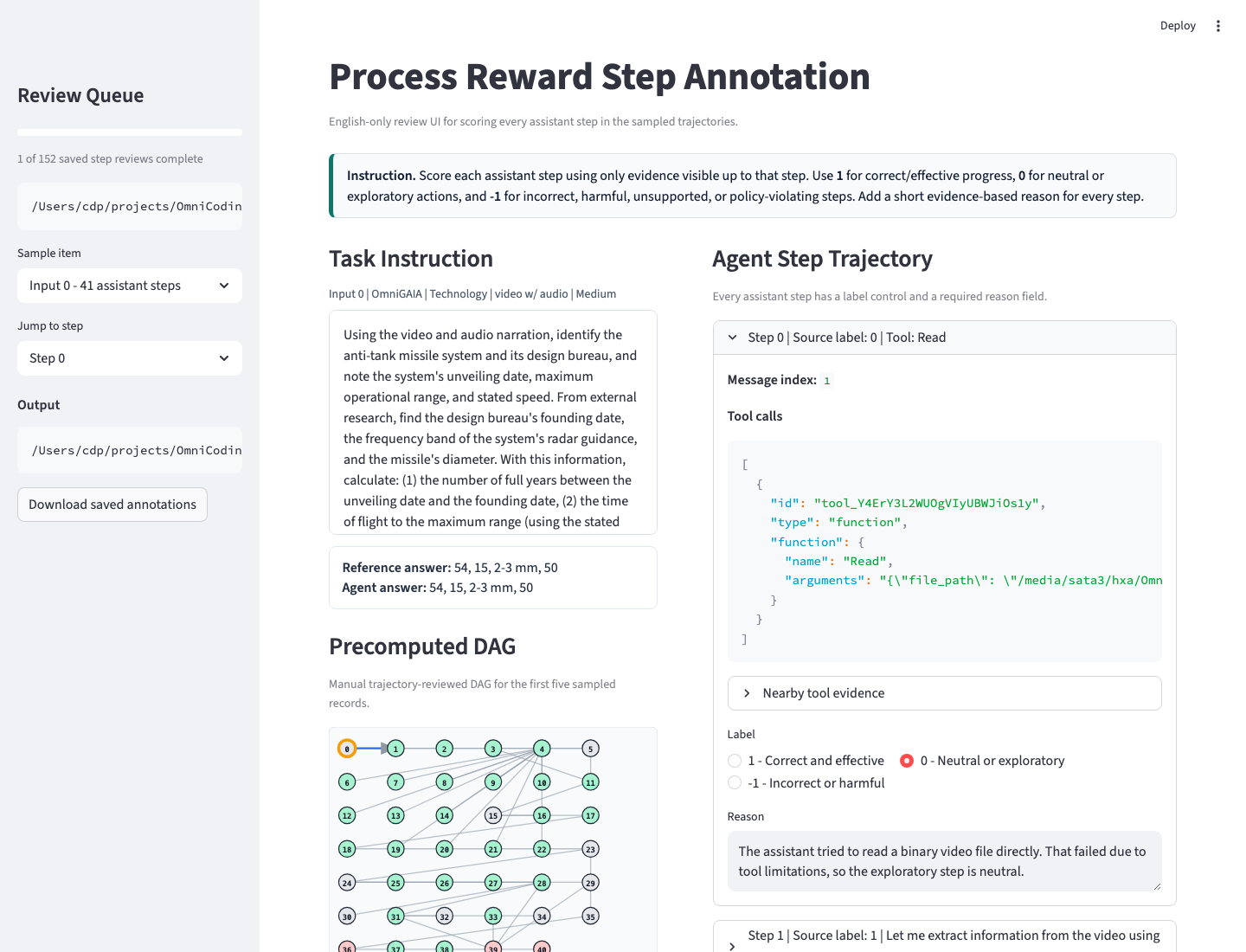}
    \caption{Annotation interface of our process-level dataset.}
    \label{fig:process_benchmark_anno}
\end{figure}

We additionally apply \textbf{error propagation}: once a step is labeled $-1$, dependent later assistant steps remain $-1$ until the agent explicitly corrects the mistake or starts an independent subtask no longer relying on the bad premise. We provide our annotation interface in Figure~\ref{fig:process_benchmark_anno}.

Table~\ref{tab:reward_model_judge} compares predictions against human labels via Step Acc.\ and Primary Acc. GPT-5.5 wins on both metrics; GPT-5.4 is second on step agreement; Qwen3.6-35B-A3B matches GPT-5.4 on error-type accuracy at much higher throughput; GPT-5.4-mini and DeepSeek-V4-Flash trail on step reward, indicating that smaller models are not yet adequate for fine-grained process supervision here. Based on these results, we adopt GPT-5.5 as the process-level judge for all large-scale failure mode analyses in the main paper.


\clearpage
\section{Implementation Details}
\label{appendix: implement details}

\subsection{\method Reward Hyperparameters}
\label{appendix:reward_hparams}

Table~\ref{tab:reward_hparams} lists the reward weights and group-level coefficients used in all RL experiments. The values are shared across the 9B and 27B runs.

\begin{table}[h]
\centering
\small
\setlength{\tabcolsep}{8pt}
\caption{Reward and group-shaping hyperparameters used in our RL training.}
\label{tab:reward_hparams}
\begin{tabular}{l c l}
\toprule
\textbf{Symbol} & \textbf{Value} & \textbf{Description} \\
\midrule
\multicolumn{3}{l}{\textit{Per-trajectory reward components}} \\
\midrule
$w_{\text{correct}}$        & $+1.0$  & Base reward for gated-correct trajectories. \\
$w_{\text{fmt}+}$           & $+0.2$  & Format bonus: wrong answer, clean \texttt{task\_complete} exit, non-empty \verb|<answer>|. \\
$w_{\text{fmt}-}$           & $-0.2$  & Format penalty: any other exit (timeout, step-limit, no answer, etc.). \\
$w_{\mathrm{mod}}$          & $-0.3$  & Modality penalty when video/audio task is solved without a corresponding tool. \\
$w_{\mathrm{tool}}$         & $0.5$   & Bad-tool weight; effective penalty is $-w_{\mathrm{tool}}\cdot P_{\text{bad}}\in[-0.5,0]$. \\
\midrule
\multicolumn{3}{l}{\textit{Group-level shaping}} \\
\midrule
$G$                         & $8$     & Sibling rollouts per task (i.e., \texttt{n\_samples\_per\_prompt}). \\
$\alpha$                    & $0.2$   & Length-penalty slope. \\
$f_{\min}$                  & $0.5$   & Minimum length factor for correct samples. \\
$\epsilon_A$                & $10^{-6}$ & Stabilizer in advantage normalization. \\
\midrule
\multicolumn{3}{l}{\textit{GSPO clipping}} \\
\midrule
$\epsilon_-$                & $0.20$  & Lower clip threshold for $\rho_i(\theta)$. \\
$\epsilon_+$                & $0.28$  & Upper clip threshold (DAPO clip-higher). \\
\bottomrule
\end{tabular}
\end{table}

\header{Tool universes for modality detection.}
Modality engagement (used by $w_{\mathrm{mod}}$ and the correctness gate) is detected by substring-matching the bash text inside \texttt{execute\_commands} arguments against the following tool universes:
\begin{itemize}[leftmargin=*,itemsep=1pt]
    \item \textbf{Video:} \texttt{ffmpeg}, \texttt{ffprobe}, \texttt{ffplay}, \texttt{yt-dlp}, \texttt{mpv}, \texttt{vlc}, \texttt{VideoCapture}, \texttt{VideoWriter}, \texttt{VideoFileClip}, \texttt{moviepy}, \texttt{decord}, \texttt{VideoReader}, \texttt{imageio}, \texttt{mediainfo}.
    \item \textbf{Audio:} \texttt{whisper}, \texttt{faster\_whisper}, \texttt{librosa}, \texttt{torchaudio}, \texttt{soundfile}, \texttt{sf.read}, \texttt{pydub}, \texttt{AudioSegment}, \texttt{wave.open}, \texttt{sox}, \texttt{demucs}, \texttt{spleeter}, \texttt{lame}.
    \item \textbf{Image:} \texttt{PIL.Image.open}, \texttt{Image.open}, \texttt{cv2.imread}, \texttt{cv2.imwrite}, \texttt{tesseract}, \texttt{pytesseract}, \texttt{convert}, \texttt{magick}, \texttt{identify}, \texttt{mogrify}, \texttt{easyocr}, \texttt{paddleocr}, \texttt{pdftotext}, \texttt{pdftoppm}, \texttt{exiftool}, \texttt{image\_read}.
\end{itemize}
A task with $\texttt{has\_video}{=}\texttt{True}$ requires at least one match from the video or image set; a task with $\texttt{has\_audio}{=}\texttt{True}$ requires at least one match from the audio set. Image-only tasks are not modality-gated.

\header{Bad-tool fail categories.}
The numerator of $P_{\text{bad}}=(n_{\text{unparseable}}+n_{\text{disallowed}}+n_{\text{escape}}+n_{\text{syntax-fail}})/(1+n_{\text{tool}})$ aggregates four categories:
\begin{itemize}[leftmargin=*,itemsep=1pt]
    \item \textbf{Unparseable:} malformed JSON in tool call, missing or non-dict \texttt{arguments}, or missing tool \texttt{name}.
    \item \textbf{Disallowed:} tool name outside the allowed set $\{\texttt{execute\_commands},\,\texttt{image\_read},\,\texttt{task\_complete}\}$, or arguments matching network primitives (\texttt{curl}, \texttt{wget}, \texttt{ssh}, \texttt{scp}, \texttt{rsync}, \texttt{nc}, \texttt{telnet}, \texttt{http(s)://}).
    \item \textbf{Escape:} arguments matching gold-answer patterns (\texttt{ground\_truth}, \texttt{answer\_key}, \texttt{gold\_answer}, \texttt{*.gold}, \texttt{all\_final.jsonl}, \texttt{rl\_train.jsonl}, \texttt{sft\_train.jsonl}), or absolute paths resolving outside the workspace. Any escape event additionally triggers a hard zero on correctness.
    \item \textbf{Syntax-fail:} the tool reply contains \texttt{SyntaxError}, \texttt{invalid syntax}, \texttt{unexpected EOF}, \texttt{parse error}, \texttt{unexpected token}, \texttt{missing operand}, or \texttt{unterminated string}.
\end{itemize}
Unparseable tool calls short-circuit the other static checks, since they never actually execute in the sandbox.

\header{Removed-trajectory criterion.}
A trajectory is marked \emph{removed} (score zeroed, sample masked from the policy-gradient loss) iff the agent emits no \verb|<answer>| \emph{and} the exit reason is one of \texttt{timeout}, \texttt{step\_limit}, or \texttt{context\_overflow}. Group-level statistics (mean, std, dynamic filter) are computed over active (non-removed) samples only.

\subsection{Prompts in Benchmark}

The exact prompts used in our experiments are listed in Figure \ref{fig:calibration_self_iteration_prompt_1}, \ref{fig:calibration_self_iteration_prompt_2}, \ref{fig:socialomni-prompt}, \ref{fig:socialomni-level2-prompt}, \ref{fig:long-audiovideo-prompt}, \ref{prompt_VideoZeroBench}.

\begin{figure}[htbp]
    \centering
\begin{tcolorbox}[enhanced,
  attach boxed title to top center={yshift=-3mm,yshifttext=-1mm},
  boxrule=0.9pt,
  colback=gray!00,
  colframe=black!50,
  colbacktitle=Gray,
  title=Prompt 1: Calibration-set self-iteration,
  boxed title style={size=small,colframe=Gray}
]

\begin{minipage}{\linewidth}
{\small
You are updating a reusable benchmark agent guide.

Read these files in the current workspace:

- \texttt{\{previous\_agent\_md\_name\}}

- \texttt{\{summary\_json\_name\}}

Task:

1. Update the previous guide into a better next-round guide.

2. Use only generic, reusable tactics suggested by the sanitized summary.

3. Do not include benchmark-specific facts, named entities, dates, exact answers, or any clues tied to individual cases.

4. Do not quote or paraphrase specific questions.

5. Keep the guide concise, operational, and directly useful during future runs.

What to extract from the summary:

- recurring failure patterns

- search-breadth problems

- weak verification habits

- answer-format mistakes

- underused or misused tools

- signals about when a workflow should escalate from local inspection to search, OCR, ASR, calculation, or multi-step verification

- patterns in required tools and tool-call usage that suggest missing playbook guidance

- the abstract difference between OmniGAIA reference answer processes and the model's observed processes

- reusable workflow lessons from cases where the reference process used stronger verification, better ordering, or better skill/tool composition than the model did
}
\end{minipage}

\end{tcolorbox}
    \caption{First part of prompt for calibration-set self-iteration.}
    \label{fig:calibration_self_iteration_prompt_1}
\end{figure}

\begin{figure}[htbp]
    \centering
\begin{tcolorbox}[enhanced,
  attach boxed title to top center={yshift=-3mm,yshifttext=-1mm},
  boxrule=0.9pt,
  colback=gray!00,
  colframe=black!50,
  colbacktitle=Gray,
  title=Prompt 2: Calibration-set self-iteration,
  boxed title style={size=small,colframe=Gray}
]

\begin{minipage}{\linewidth}
{\small
Additional requirements for the new guide:

- explicitly state that the agent must not access, read, inspect, rely on, or search for any content outside the current workspace

- include a short section that summarizes when to use specific skills or capability families

- include a short section that summarizes tool-use heuristics

- include a short section that summarizes better end-to-end answering workflows derived from reference-process-vs-model-process comparisons

- for tool use, prioritize generic guidance such as:
  - when \texttt{web\_search} should be preferred over guesswork
  - when shell inspection should precede search
  - when OCR / ASR / frame extraction / calculation should be combined
  - when to stop searching and verify the answer format instead

- when the summary provides reference-process signals, compare them against the model process and convert the differences into better workflow steps or skill-selection rules

- make the guide explain not just which tool to use, but in what order and with what verification checkpoints

- if the reference process suggests a stronger answer pipeline, rewrite that into a reusable skill/playbook description rather than a case-specific tip

- if the summary suggests over-searching or under-searching, reflect that explicitly in the updated guide

- if the summary suggests tools were used but not verified against the final answer, add a guardrail for that

- if the summary suggests repeated empty or failed predictions, add concise recovery rules

Required output:

- return markdown only

- start with \texttt{\# Agent Guide}

- prefer short sections and flat bullet lists

- keep the guide high-level enough to generalize, but concrete enough to change behavior

Focus areas:

- search strategy

- tool selection

- skill / capability-family selection

- workflow / process design

- verification discipline

- common failure guards

- answer-format discipline

Hard prohibitions:

- no benchmark-specific examples

- no copied phrases from individual questions

- no named entities, dates, locations, titles, or answer strings from the sampled cases

- no hidden hints that would advantage the exact benchmark subset
}
\end{minipage}

\end{tcolorbox}
    \caption{Second part of prompt for calibration-set self-iteration.}
    \label{fig:calibration_self_iteration_prompt_2}
\end{figure}


\begin{figure}[htbp]
    \centering
\begin{tcolorbox}[enhanced,attach boxed title to top center={yshift=-3mm,yshifttext=-1mm},boxrule=0.9pt, 
  colback=gray!00,colframe=black!50,colbacktitle=Gray,
  title=Prompt for testing SocialOmni Leve1,
  boxed title style={size=small,colframe=Gray} ]

\begin{minipage}{\linewidth}
{\small
You are evaluating a SocialOmni benchmark item in an isolated local workspace.

Work only inside the current workspace.

Your current working directory is the workspace root for this run.

Treat the current working directory tree as the only allowed workspace root.

Do not access, read, inspect, rely on, or search for files, directories, or other local content outside the current workspace.

Never open absolute paths or parent-directory paths outside the workspace, and never use symlinks, environment discovery, shell expansion, or search commands to reach files beyond the workspace root.

Treat any host file, cache, home-directory content, sibling directory, mounted dataset source, or path outside this workspace as strictly forbidden, even if it appears readable from the shell.

Use only the staged files listed below; do not assume any hidden benchmark files exist.

You may install dependencies without sudo when necessary, but keep them inside the current workspace.

User-level install prefixes and caches are redirected into the workspace for this run.

Available staged files:

\texttt{[LOCAL\_VIDEO\_PATH]}

Network access may be available for shell commands in this run.

You may use the network only when it materially helps process the current sample, including downloading tools or packages into the current workspace only.

Do not use the network to search for benchmark answers, leaked annotations, dataset-specific solutions, or existing evaluation outputs.

\texttt{[SYSTEM]}

\texttt{[LEVEL1\_SYSTEM\_PROMPT]}

\texttt{[LOCAL\_VIDEO]}

Analyze the local video file at: \texttt{[LOCAL\_VIDEO\_PATH]}

\texttt{[ASR]}

\texttt{[ASR\_CONTENT\_IF\_ENABLED]}

Options:

\texttt{A. ...}

\texttt{B. ...}

\texttt{C. ...}

\texttt{D. ...}

\texttt{[LEVEL1\_USER\_PROMPT\_BASE]}

\texttt{[ANSWER\_FORMAT]}

Output requirements:

Return ONLY the single option letter (A, B, C, D).

Do not repeat the option text.

Do not provide any explanation, reasoning, analysis, punctuation, markdown, or extra words.

Your entire response must be exactly one of: \texttt{A/B/C/D}.

\texttt{[QUESTION\_TEXT]}
}
\end{minipage}

\end{tcolorbox}
    \caption{Prompt for testing SocialOmni Level1.}
    \label{fig:socialomni-prompt}
\end{figure}


\begin{figure}[htbp]
    \centering
\begin{tcolorbox}[enhanced,attach boxed title to top center={yshift=-3mm,yshifttext=-1mm},boxrule=0.9pt, 
  colback=gray!00,colframe=black!50,colbacktitle=Gray,
  title=Prompt for testing SocialOmni Level2,
  boxed title style={size=small,colframe=Gray} ]

\begin{minipage}{\linewidth}
{\small
You are evaluating a SocialOmni benchmark item in an isolated local workspace.

Work only inside the current workspace.

Your current working directory is the workspace root for this run.

Treat the current working directory tree as the only allowed workspace root.

Do not access, read, inspect, rely on, or search for files, directories, or other local content outside the current workspace.

Never open absolute paths or parent-directory paths outside the workspace, and never use symlinks, environment discovery, shell expansion, or search commands to reach files beyond the workspace root.

Treat any host file, cache, home-directory content, sibling directory, mounted dataset source, or path outside this workspace as strictly forbidden, even if it appears readable from the shell.

Use only the staged files listed below; do not assume any hidden benchmark files exist.

You may install dependencies without sudo when necessary, but keep them inside the current workspace.

User-level install prefixes and caches are redirected into the workspace for this run.

Available staged files:

\texttt{[LOCAL\_PREFIX\_VIDEO\_PATH]}

Network access may be available for shell commands in this run.

You may use the network only when it materially helps process the current sample, including downloading tools or packages into the current workspace only.

Do not use the network to search for benchmark answers, leaked annotations, dataset-specific solutions, or existing evaluation outputs.

\texttt{[SYSTEM]}

\texttt{[LEVEL2\_SYSTEM\_PROMPT]}

\texttt{[LOCAL\_VIDEO]}

Analyze the local prefix video file at: \texttt{[LOCAL\_PREFIX\_VIDEO\_PATH]}

\texttt{[ASR]}

\texttt{[FULL\_ASR\_IF\_ENABLED]}

\texttt{[LEVEL2\_USER\_PROMPT\_BASE]}

Task 1 (Q1):

Question: \texttt{[Q1\_QUESTION\_TEXT]}

Options:

\texttt{A. [OPTION\_A]}

\texttt{B. [OPTION\_B]}

Task 2 (Q2):

Question: \texttt{[Q2\_QUESTION\_TEXT]}

If your Q1 answer is A, provide a concise natural utterance for Q2.

If your Q1 answer is B, set Q2 to N/A.

Output requirements:

Do not provide explanation, reasoning, analysis, markdown, or extra words.

Return exactly in the following format:

\texttt{<q1>A\_or\_B</q1>}

\texttt{<q2>utterance\_or\_NA</q2>}
}
\end{minipage}

\end{tcolorbox}
    \caption{Prompt for testing SocialOmni Level2.}
    \label{fig:socialomni-level2-prompt}
\end{figure}


\begin{figure}[htbp]
    \centering
\begin{tcolorbox}[enhanced,attach boxed title to top center={yshift=-3mm,yshifttext=-1mm},boxrule=0.9pt, 
  colback=gray!00,colframe=black!50,colbacktitle=Gray,
  title=Prompt for testing LVOmniBench,
  boxed title style={size=small,colframe=Gray} ]

\begin{minipage}{\linewidth}
{\small
You are evaluating a long audio-video benchmark item in an isolated local workspace.

Work only inside the current workspace.

Do not access, read, inspect, rely on, or search for any files, directories, or other content outside the current workspace.

Never open absolute paths or parent-directory paths outside the workspace, and never use symlinks, environment discovery, shell expansion, or search commands to reach files beyond the workspace root.

Treat any host file, cache, home-directory content, sibling directory, mounted dataset source, or path outside the current workspace as strictly forbidden, even if it appears readable from the shell.

Use the staged video file(s) listed below and any local tools available in the workspace.

Do not assume access to any original benchmark directories outside this workspace.

Network access may be available for shell commands in this run.

You may use the network only when it materially helps process the current sample, including downloading tools or packages into the current workspace only.

Do not use the network to search for benchmark answers, leaked annotations, dataset-specific solutions, or existing evaluation outputs.

Available staged files:

\texttt{[STAGED\_VIDEO\_FILES]}

Use the benchmark's official answering format below.

Question: \texttt{[QUESTION\_TEXT]}

Options:

\texttt{A. ...}

\texttt{B. ...}

\texttt{C. ...}

\texttt{D. ...}

Select the best answer from the options above. Directly provide the letter representing your choice (A/B/C/D) and nothing else. Do not include the full text of the option, do not provide any explanation.
}
\end{minipage}

\end{tcolorbox}
    \caption{Prompt for testing LVOmniBench.}
    \label{fig:long-audiovideo-prompt}
\end{figure}


\begin{figure}[htbp]
    \centering
\begin{tcolorbox}[enhanced,attach boxed title to top center={yshift=-3mm,yshifttext=-1mm},boxrule=0.9pt, 
  colback=gray!00,colframe=black!50,colbacktitle=Gray,
  title=Prompt for testing VideoZeroBench,
  boxed title style={size=small,colframe=Gray} ]

\begin{minipage}{\linewidth}
{\small
You are evaluating VideoZeroBench questions in a sandboxed local workspace.

The model can access only the staged video file and files it creates during the current run. Do not access, read, inspect, rely on, or search for files outside the sandbox, including benchmark annotations, gold answers, evidence windows, evidence boxes, previous evaluation outputs, cached results, or any other files that may reveal the answer.

Network access is allowed only for general-purpose assistance that does not reveal benchmark-specific information. Do not use the network to search for the benchmark name, dataset, question text, video id, video filename, leaked annotations, gold answers, existing evaluation outputs, or papers/pages that evaluate or discuss the same sample. Do not retrieve any content that directly or indirectly reveals the answer.

If video inspection is needed, use information derived only from the provided staged video and the question. Do not use external sources or prior knowledge to retrieve or infer benchmark-specific answers.

All questions below refer to the same video and must be answered in a single response.

Available staged video:

- \texttt{\{VIDEO\_FILENAME\}.mp4}

Video id / filename: \texttt{\{VIDEO\_FILENAME\}.mp4}

Requested evaluation level: Level-3.

Return ONLY valid JSON.

Return a JSON array with exactly one object per \texttt{question\_id} shown below.

Do not include markdown fences, explanations, extra keys, chain-of-thought, or commentary.

The entire final response must be the JSON array itself.

Your first output character must be \texttt{[} and your final output character must be \texttt{]}.

If you cannot determine an answer, use an empty string.

Required output object schema:

\texttt{\{}\\
\hspace*{1.5em}\texttt{"question\_id": "string",}\\
\hspace*{1.5em}\texttt{"level3\_answer": "string"}\\
\texttt{\}}

Question bundle:

\texttt{[}\\
\hspace*{1.5em}\texttt{\{}\\
\hspace*{3em}\texttt{"question\_id": "\{QUESTION\_ID\_1\}",}\\
\hspace*{3em}\texttt{"question": "\{QUESTION\_TEXT\_1\}"}\\
\hspace*{1.5em}\texttt{\},}\\
\hspace*{1.5em}\texttt{\{}\\
\hspace*{3em}\texttt{"question\_id": "\{QUESTION\_ID\_2\}",}\\
\hspace*{3em}\texttt{"question": "\{QUESTION\_TEXT\_2\}"}\\
\hspace*{1.5em}\texttt{\}}\\
\texttt{]}

Level-3 QA rule:

- This is a question answering task without benchmark-provided temporal or spatial evidence.

- Answer according to the video content and the question.

- Return only the final answer text in \texttt{level3\_answer}.
}
\end{minipage}

\end{tcolorbox}
    \caption{Prompt for testing VideoZeroBench, where all questions for the same video are answered together in a single response.}
    \label{prompt_VideoZeroBench}
\end{figure}

\subsection{Experimental Configuration Details}
\label{app:exp-config-details}

This subsection summarizes the evaluation configurations used in our experiments for each agent and omnimodal model setting, including controller settings, model assignments, modality-processing rules, and benchmark-specific runtime defaults.

\subsubsection{Agent-Omni}

\header{Controller and graph configuration.}
Agent-Omni is evaluated with a fixed multi-agent graph consisting of modality summarization, master reasoning, modality-specific execution, and master decision. The default reasoning budget is two rounds, i.e., \texttt{max\_round\_num=2}. The global retry budget is \texttt{retry\_times=10}.

\header{Model assignment.}
The committed benchmark configuration uses the following model split. \textbf{We use even stronger models compared to the configuration in their paper}:
\begin{itemize}[itemsep=0pt, leftmargin=*]
  \item orchestrate agent: \texttt{openai/gpt-5.4};
  \item text agent: \texttt{openai/gpt-5.4};
  \item image agent: \texttt{openai/gpt-5.4};
  \item video agent: local \texttt{Qwen/Qwen2.5-Omni-7B};
  \item audio agent: local \texttt{Qwen/Qwen2.5-Omni-7B}.
\end{itemize}

\header{Generation settings.}
The orchestrate, text, and image agents use \texttt{max\_tokens=4096} and \texttt{temperature=0}. The video agent uses \texttt{max\_tokens=2048}, \texttt{temperature=0}, \texttt{max\_image\_input=12}, \texttt{max\_frames\_num=448}, and \texttt{image\_size=448 $\times$ 448}. The audio agent uses \texttt{max\_tokens=4096} and \texttt{temperature=0}.

\header{Modality processing.}
For video benchmarks, the audio track is extracted from each video and provided jointly with the original video input, so the effective input to the graph is audio-video rather than video only. All four modality branches remain part of the graph, while the master reasoning module determines which agents receive non-empty questions.

\header{Output normalization.}
For multiple-choice evaluation, the graph output is not used directly. Instead, a separate selector or finalizer model is applied after graph execution. By default, this selector/finalizer uses the same model as the master agent.

\header{LVOmniBench configuration.}
The default runtime configuration is:
\begin{itemize}[itemsep=0pt, leftmargin=*]
  \item \texttt{concurrent\_limit=1},
  \item \texttt{max\_round\_num=2},
  \item \texttt{request\_timeout=600},
  \item \texttt{ffmpeg\_timeout=300},
  \item \texttt{empty\_answer\_retry\_rounds=3}.
\end{itemize}

\header{SocialOmni configuration.}
The default runtime configuration is:
\begin{itemize}[itemsep=0pt, leftmargin=*]
  \item \texttt{max\_concurrency=1},
  \item \texttt{max\_round\_num=2},
  \item \texttt{request\_timeout=600},
  \item \texttt{ffmpeg\_timeout=300},
  \item \texttt{empty\_answer\_retry\_rounds=3}.
\end{itemize}
For Level-2 evaluation, a prefix video is first cropped up to the target timestamp. The model predicts Q1 before Q2, and Q2 is executed only when the gold Q1 label is \texttt{A} and the Q1 prediction is correct. The default judge model for Level-2 is \texttt{gpt-5.4}.

\header{VideoZeroBench configuration.}
The default runtime configuration is:
\begin{itemize}[itemsep=0pt, leftmargin=*]
  \item \texttt{level=level3},
  \item \texttt{group\_by\_video=True},
  \item \texttt{concurrent\_limit=1},
  \item \texttt{max\_round\_num=2},
  \item \texttt{request\_timeout=600},
  \item \texttt{ffmpeg\_timeout=300},
  \item \texttt{empty\_answer\_retry\_rounds=2}.
\end{itemize}
When grouped by video, all questions attached to the same video are answered from a single graph run.

\subsubsection{OmniAgent}

\header{Controller configuration.}
OmniAgent is evaluated as a single tool-using controller. The default controller budget is \texttt{max\_iterations=30}. The controller model uses \texttt{temperature=1}, and the default reasoning effort is set to \texttt{high} when supported by the serving API.

\header{Model assignment.}
The default released configuration is listed below. \textbf{We use even stronger models compared to the configuration in their paper}:
\begin{itemize}[itemsep=0pt, leftmargin=*]
  \item controller model: \texttt{openai/gpt-5.4};
  \item video tool model: local \texttt{Qwen/Qwen2.5-Omni-7B};
  \item audio tool model: \texttt{openai/gpt-4o-audio-preview}.
\end{itemize}

\header{Tool set.}
The default core tools are:
\begin{itemize}[itemsep=0pt, leftmargin=*]
  \item \texttt{audio\_global\_caption},
  \item \texttt{audio\_qa},
  \item \texttt{audio\_ASR},
  \item \texttt{Audio\_EventList},
  \item \texttt{Audio\_EventLocation},
  \item \texttt{video\_global\_qa},
  \item \texttt{video\_clip\_qa},
  \item \texttt{video\_metadata}.
\end{itemize}
External web and code tools are implemented but disabled by default in the benchmark runs.

\header{Video sampling configuration.}
The default multimodal settings are:
\begin{itemize}[itemsep=0pt, leftmargin=*]
  \item \texttt{VIDEO\_INPUT\_MODE=frames},
  \item \texttt{VIDEO\_TARGET\_FPS=0.5},
  \item \texttt{VIDEO\_CLIP\_TARGET\_FPS=2.0},
  \item \texttt{VIDEO\_MAX\_FRAMES=24},
  \item \texttt{VIDEO\_CLIP\_MAX\_FRAMES=40},
  \item \texttt{VIDEO\_MAX\_EDGE=1024},
  \item \texttt{MULTIMODAL\_MAX\_TOKENS=4096}.
\end{itemize}
For oversized or empty multimodal responses, the video tool falls back to smaller payloads using \texttt{16/12/8} frames and \texttt{768/640/512} maximum edge size.

\header{Audio processing.}
The audio tool can consume audio files directly and can also extract audio from video inputs. Audio event localization and timeline generation are performed with chunk-based prompting followed by a merge step.

\header{LVOmniBench configuration.}
The default runtime configuration is:
\begin{itemize}[itemsep=0pt, leftmargin=*]
  \item \texttt{concurrent\_limit=1},
  \item \texttt{max\_iterations=30},
  \item \texttt{item\_timeout=5400},
  \item \texttt{empty\_answer\_retry\_rounds=3},
  \item external tools disabled by default.
\end{itemize}

\header{SocialOmni configuration.}
The default runtime configuration is:
\begin{itemize}[itemsep=0pt, leftmargin=*]
  \item \texttt{concurrent\_limit=2},
  \item \texttt{max\_iterations=30},
  \item \texttt{item\_timeout=5400},
  \item \texttt{empty\_answer\_retry\_rounds=3},
  \item external tools disabled by default.
\end{itemize}
For Level-2 evaluation, a prefix video is cached up to the target timestamp. Q1 is executed before Q2, and Q2 is run only when the gold Q1 label is \texttt{A} and the Q1 prediction is correct. When available, the benchmark ASR field is injected into the prompt. The Level-2 judge uses the evaluation endpoint specified by the evaluation model settings, which default to the controller-model endpoint unless overridden.

\header{VideoZeroBench configuration.}
The default runtime configuration is:
\begin{itemize}[itemsep=0pt, leftmargin=*]
  \item \texttt{level=level3},
  \item \texttt{request\_unit=video},
  \item \texttt{concurrent\_limit=1},
  \item \texttt{max\_iterations=30},
  \item \texttt{item\_timeout=5400},
  \item \texttt{empty\_answer\_retry\_rounds=3},
  \item external tools disabled by default.
\end{itemize}
Under \texttt{request\_unit=video}, all questions corresponding to the same video are answered in one batched response, and the final answers are parsed from a JSON answer map.

\section{Licenses and Use Terms of Existing Assets}
\label{app:licenses}

We list below all third-party benchmarks, datasets, models, and software libraries used in this work, together with their licenses to the best of our knowledge. All assets are used in a manner consistent with their stated terms; we used them only for non-commercial academic research.

\subsection*{Benchmarks for Evaluation}
\begin{itemize}[leftmargin=*, itemsep=0pt]
    \item \textbf{OmniGAIA} \citep{li2026omnigaia}: an open-ended, tool-mediated factual and multi-hop omni-modal reasoning benchmark. Released under the MIT License for the code and CC-BY-4.0 for the data annotations; the underlying media inherits the licenses of its original public sources.
    \item \textbf{SocialOmni} \citep{xie2026socialomni}: an audio-visual social interaction benchmark for omni models. Released under the MIT License for the code; data is provided for non-commercial research use, with the underlying videos governed by their original platform licenses.
    \item \textbf{LVOmniBench} \citep{tao2026lvomnibench}: a long audio-video understanding benchmark. Released under the MIT License for the code and Apache-2.0 for the released annotations; underlying videos retain their original licenses.
    \item \textbf{VideoZeroBench} \citep{meng2026videozerobench}: a video MLLM benchmark with spatio-temporal evidence verification. Released under the MIT License for the code; data annotations are provided under CC-BY-4.0 for academic use.
\end{itemize}

\subsection*{Datasets Used in \textsc{OmniCoding}}
\begin{itemize}[leftmargin=*, itemsep=0pt]
    \item \textbf{OmniGAIA-SFT-2K} \citep{li2026omnigaia}: the supervised-trajectory subset distributed with OmniGAIA. Released alongside OmniGAIA under the same MIT/CC-BY-4.0 terms.
    \item \textbf{OmniVideoBench} \citep{li2025omnivideobench}: an audio-visual understanding benchmark for omni MLLMs. Released under the MIT License; underlying videos retain their original public licenses.
    \item \textbf{AVUTBenchmark} \citep{yang2025avut}: an audio-centric video understanding benchmark designed to remove text shortcuts. Released under the MIT License for code and CC-BY-4.0 for annotations.
    \item \textbf{Video-MME-v2} (audio-visual / temporal subset) \citep{fu2026videommev2}: a comprehensive video understanding benchmark. Released under the MIT License (code) and CC-BY-NC-4.0 (data) for academic, non-commercial use.
\end{itemize}

\subsection*{Source Material for \textsc{TerminalBench-O}}
All raw media in \textsc{TerminalBench-O} is drawn from authentic public sources:
\begin{itemize}[leftmargin=*, itemsep=0pt]
    \item \textbf{Internet Archive}: content used under the licenses indicated on each Internet Archive item page (typically Public Domain, CC-BY, or CC0).
    \item \textbf{HuggingFace public datasets}: each used under the dataset-specific license declared on its HuggingFace card (the specific licenses are listed in our benchmark release card).
    \item \textbf{Public web videos}: incorporated only when the source platform's terms permit research use, with attribution preserved in our task metadata.
\end{itemize}

\subsection*{Models}
\begin{itemize}[leftmargin=*, itemsep=0pt]
    \item \textbf{Qwen-3.5-9B} and \textbf{Qwen-3.6-27B} \citep{qwen35Omni}: open-weight base models we fine-tune. Released by Alibaba under the Tongyi Qianwen / Apache-2.0-style license, which permits research and commercial use subject to the model's published terms.
    \item \textbf{Whisper (large-v3)}: ASR model used inside agent rollouts. Released by OpenAI under the MIT License.
    \item \textbf{Closed-source models} (GPT-5.4, Claude Opus 4.6 / Sonnet 4.6, Gemini 3.1 Pro, Gemini 3 Flash, etc.) are accessed exclusively through their official APIs in compliance with each provider's terms of service.
\end{itemize}

\subsection*{Software and Libraries}
\begin{itemize}[leftmargin=*, itemsep=0pt]
    \item \textbf{ffmpeg / ffprobe}: LGPL-2.1+ (with GPL components in some builds).
    \item \textbf{Tesseract OCR}: Apache-2.0.
    \item \textbf{OpenCV}: Apache-2.0.
    \item \textbf{Librosa}: ISC License.
    \item \textbf{ImageMagick}: ImageMagick License (Apache-2.0-compatible).
    \item \textbf{yt-dlp}: Unlicense (public domain).
    \item \textbf{Python} and standard scientific stack (NumPy, SciPy, etc.): each under their respective permissive licenses (BSD/PSF/MIT).
\end{itemize}

\subsection{Omni-terminal Harness}
\label{app:omnicoding-harness}

This appendix details our terminal-agent harness, adapted from Terminus-KIRA~\citep{terminuskira2026} and designed around a single operation: turning raw multimodal files into small, checkable evidence. Each episode begins with an isolated workspace containing the question and the staged media files. Rather than packing media into the model context, files remain on disk, and the agent must decide how to inspect them: extracting audio, sampling frames, running OCR or ASR, reading metadata, writing scripts, and verifying intermediate results before answering. The agent operates through a small action space (\texttt{execute\_commands}, \texttt{image\_read}, \texttt{task\_complete}) and expresses higher-level media operations through code and command-line tools (\texttt{ffmpeg}/\texttt{ffprobe}, ASR, OCR, Python, OpenCV, NumPy, and self-authored scripts) rather than benchmark-specific APIs. Beyond execution, the harness is centered on trajectory collection: every tool call, terminal observation, image observation, final answer, exit reason, and reward field is serialized so the same rollout can be reused for SFT filtering, RL replay, and failure analysis.

Concretely, the harness has four properties that are important for training:
\begin{itemize}[leftmargin=*,nosep]
    \item \textbf{Filesystem-grounded media processing.} Raw video, audio, images, and documents are workspace state. The agent learns to transform them into transcripts, key frames, OCR text, timestamps, tables, and other compact evidence.
    \item \textbf{Persistent terminal interaction.} Commands run in a stateful bash session with marker-based completion, bounded observations, timeout recovery, and support for scripts written during the episode.
    \item \textbf{Replay-faithful multimodal observations.} \texttt{image\_read} inserts the actual image into the main conversation, rather than replacing it with a caption from a separate model. The saved rollout therefore matches the observation distribution used for multimodal SFT/RL.
    \item \textbf{Qwen-compatible rollout collection.} The tool schema is flat, malformed tool-call blocks are conservatively recovered, and requests are routed through sticky SGLang endpoints to preserve prefix-cache locality during large parallel rollouts.
\end{itemize}

\header{Episode state.}
Each task starts from an isolated workspace
\[
s_0 = (q, W_0, U_0),
\]
where \(q\) is the task instruction, \(W_0\) contains the staged input files, and \(U_0\) contains the available command-line tools and Python libraries. 
The model does not receive the raw video or audio stream as context. 
Instead, it interacts with files through the terminal. 
During a rollout, the workspace may accumulate derived artifacts such as extracted frames, audio clips, ASR transcripts, OCR outputs, metadata files, temporary scripts, and numerical tables. 
The final prediction is accepted only from the benchmark-normalized \texttt{\textless answer\textgreater...\textless/answer\textgreater} wrapper.

\header{Tool interface.}
The model sees three tools. 
The small tool set is deliberate: the terminal is the compositional interface, and the agent decides how to combine existing utilities with code it writes during the episode.

\begin{table}[t]
\centering
\small
\begin{tabular}{@{}p{0.24\linewidth}p{0.68\linewidth}@{}}
\toprule
Tool & Function \\
\midrule
\texttt{execute\_commands} &
Runs one shell command in a persistent bash session. 
The schema is flat: the call contains \texttt{analysis}, \texttt{plan}, and \texttt{keystrokes}. 
Multi-step work is expressed with shell syntax, scripts, or heredocs. \\
\texttt{image\_read} &
Loads an image file as a native multimodal observation for the main agent. 
This is used for frames, crops, figures, screenshots, OCR targets, and visual verification. \\
\texttt{task\_complete} &
Ends the rollout after the model has emitted the final answer wrapper. 
If the wrapper is missing, the harness issues one format reminder before scoring the run as empty. \\
\bottomrule
\end{tabular}
\caption{Model-visible tools in the \textsc{OmniCoding} harness.}
\label{tab:omnicoding-harness-tools}
\end{table}

\header{Agent loop.}
At each step, the harness sends the current conversation and tool schema to the model. 
If the model calls \texttt{execute\_commands}, the command is executed in the persistent shell and the bounded output is returned as a tool observation. 
If it calls \texttt{image\_read}, the harness first records a short tool acknowledgement and then appends the image itself as a multimodal observation in the main conversation. 
If it calls \texttt{task\_complete}, the harness checks whether a wrapped answer already appears in the trajectory. 
The loop stops on successful completion, step limit, no-tool-call exhaustion, or runtime error.

\header{Shell runtime.}
The shell is a persistent bash subprocess. 
This preserves state across turns, including the current directory, environment variables, installed packages, and files created by the agent. 
Command completion is detected by appending a hidden marker after each command and polling until the marker appears. 
Terminal observations are capped with head--tail truncation, which keeps both the beginning of long outputs and the end where errors or final answers often appear. 
When a command times out, the harness terminates the direct child processes while keeping the shell alive. 
If the shell itself dies, it is restarted and the next observation warns the model that shell-local state has been lost.

\header{Native multimodal replay.}
The main change from a standard terminal harness is the native \texttt{image\_read} path. 
A caption-only image reader is acceptable for evaluation, but it is not acceptable for multimodal policy training: the saved trajectory would contain text produced by a separate model rather than the pixels observed during rollout. 
Our harness instead stores image observations as multimodal content blocks in the main conversation. 
For Qwen-style chat templates, these messages are folded into the provider-specific tool-response format only at send time. 
The saved artifact remains provider-neutral and can be replayed for SFT or RL without changing the observation sequence.

\header{Qwen serving and tool-call robustness.}
Open-weight models served through SGLang can emit partially malformed tool-call blocks, especially when shell commands contain quotes, heredocs, or escape characters. 
We therefore use a flat \texttt{execute\_commands} schema and a conservative recovery layer. 
If the backend returns no structured tool call but the assistant text contains a recognizable \texttt{\textless tool\_call\textgreater} block, the harness reconstructs the call only when the function name is in the declared tool list and the arguments can be parsed without inventing content. 
Recovered calls are marked in the trajectory, so downstream data filtering can keep or drop them explicitly.

\header{Trajectory artifacts.}
Each rollout stores the full message sequence, parsed tool calls, shell outputs, image-observation references, final extracted answer, exit reason, and evaluation metadata. 
The same artifact supports three downstream uses. 
First, successful trajectories can be filtered into SFT examples. 
Second, RL training can replay the same observation distribution seen during rollout. 
Third, failures can be inspected at the process level: whether the agent sampled the wrong frames, trusted a bad ASR transcript, stopped too early, chose the wrong external fact, or made a calculation error.

\begin{table}[t]
\centering
\caption{Trajectory fields saved by the harness.}
\label{tab:omnicoding-trajectory-fields}
\small
\begin{tabular}{@{}p{0.28\linewidth}p{0.64\linewidth}@{}}
\toprule[1.5pt]
Artifact field & Purpose \\
\midrule
Messages & Replayable conversation with tool calls and observations. \\
Workspace outputs & Derived files such as frames, transcripts, OCR text, scripts, and tables. \\
Image references & Native multimodal payloads used by \texttt{image\_read}. \\
Answer field & Extracted text inside \texttt{\textless answer\textgreater...\textless/answer\textgreater}. \\
Exit reason & Completion, step limit, no-tool-call exit, or runtime error. \\
Reward metadata & Automatic verification result and benchmark-specific scoring fields. \\
\bottomrule[1.5pt]
\end{tabular}
\end{table}

\begin{figure*}[t]
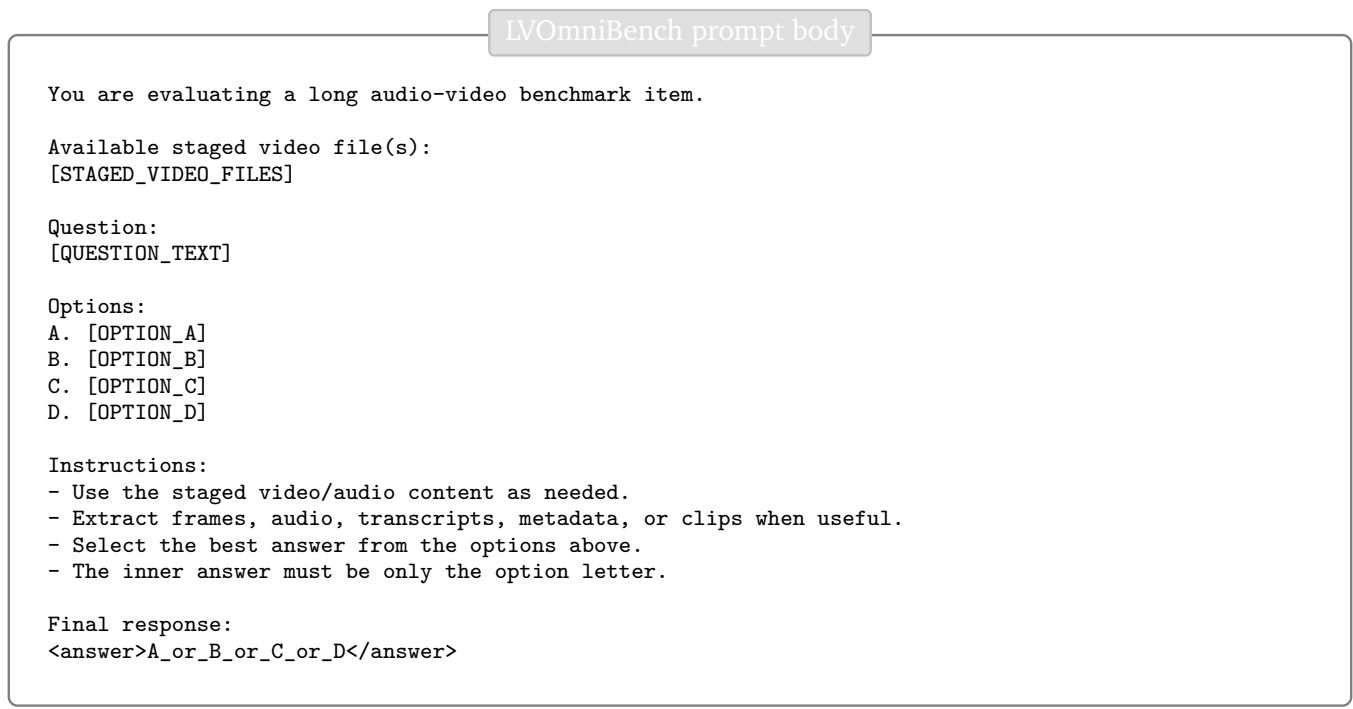

\centering
\begin{agentpromptbox}{LVOmniBench prompt body}
You are evaluating a long audio-video benchmark item.

Available staged video file(s):
[STAGED_VIDEO_FILES]

Question:
[QUESTION_TEXT]

Options:
A. [OPTION_A]
B. [OPTION_B]
C. [OPTION_C]
D. [OPTION_D]

Instructions:
- Use the staged video/audio content as needed.
- Extract frames, audio, transcripts, metadata, or clips when useful.
- Select the best answer from the options above.
- The inner answer must be only the option letter.

Final response:
<answer>A_or_B_or_C_or_D</answer>
\end{agentpromptbox}
\caption{Benchmark-specific prompt body for LVOmniBench.}
\label{fig:omnicoding-lvobench-prompt}
\end{figure*}

\header{Prompt organization.}
The prompt is factored into reusable pieces. A static system prompt defines the tool contract. A shared workspace prefix defines sandbox, network, and answer-format rules. Benchmark-specific prompt bodies supply the actual question, options, files, and output schema. Runtime reminders are used only when the model stops without a tool call or calls \texttt{task\_complete} before producing a wrapped answer. The exact prompt templates are shown in Figures~\ref{fig:omnicoding-common-prompt}--\ref{fig:omnicoding-skill-update-prompt}.

\begin{figure*}[t]
\centering
\begin{agentpromptbox}{Common workspace and answer-format prefix}
You are evaluating a benchmark item in an isolated local workspace.

Workspace rules:
- Work only inside the current workspace.
- Treat the current directory tree as the only allowed workspace root.
- Do not access, read, inspect, rely on, or search for files outside it.
- Do not open absolute paths or parent-directory paths outside the workspace.
- Do not use symlinks, shell expansion, environment discovery, or search
  commands to reach host files, caches, sibling directories, mounted datasets,
  benchmark annotations, gold answers, evidence windows, or prior outputs.
- Use only the staged files listed below and files you create during this run.

Network rules:
- Network access may be available.
- Use it only when it materially helps process the current sample, such as
  installing a dependency into the workspace or checking general public facts.
- Do not search for benchmark names, question text, video ids, file names,
  leaked annotations, gold answers, existing evaluation outputs, or pages that
  directly reveal the answer to this sample.

Available staged files:
[STAGED_FILES]

Question:
[QUESTION_TEXT]

Final-answer protocol:
- First determine the benchmark-specific answer text.
- The inner answer must follow the benchmark's official format.
- Emit the final prediction exactly as:
  <answer>FINAL_ANSWER</answer>
- Do not include reasoning, markdown, or extra text inside the answer wrapper.
\end{agentpromptbox}
\caption{Shared workspace, leakage-prevention, network-use, and final-answer rules.}
\label{fig:omnicoding-common-prompt}
\end{figure*}

\begin{figure*}[t]
\centering
\begin{agentpromptbox}{Runtime reminder prompts}
[No-tool-call reminder]

Please continue solving the task.
If you need more evidence, run a tool.
If you have enough evidence, emit the final answer using the exact answer
wrapper required by the task:
<answer>FINAL_ANSWER</answer>

Unwrapped prose is treated as no answer.
Do not stop, do not call task_complete, and do not respond with plain text
unless it is the wrapped final answer.

[Missing-answer reminder after task_complete]

task_complete was called, but no <answer>...</answer> wrapper is present in
the trajectory. The grader extracts the prediction only from that wrapper.

In your next assistant turn, emit exactly:
<answer>YOUR_ANSWER</answer>

After that answer appears in the trajectory, call task_complete in the next
turn. Prefer a plain assistant answer over shell echo, because shell quoting
can corrupt answers containing apostrophes, angle brackets, or dollar signs.
\end{agentpromptbox}
\caption{Runtime reminders used when the model stops without a tool call or calls \texttt{task\_complete} before producing a wrapped answer.}
\label{fig:omnicoding-reminder-prompt}
\end{figure*}

\begin{figure*}[t]
\centering
\begin{agentpromptbox}{SocialOmni prompt bodies}
[SocialOmni Level 1]

Analyze the local video file:
[LOCAL_VIDEO_PATH]

Optional ASR transcript, if provided by the benchmark:
[ASR_CONTENT]

Options:
A. [OPTION_A]
B. [OPTION_B]
C. [OPTION_C]
D. [OPTION_D]

Question:
[QUESTION_TEXT]

Output:
Return the single best option letter as the inner answer.
The final response must be:
<answer>A_or_B_or_C_or_D</answer>

[SocialOmni Level 2]

Analyze the local prefix video file:
[LOCAL_PREFIX_VIDEO_PATH]

Optional full ASR transcript, if provided by the benchmark:
[FULL_ASR_CONTENT]

Task 1 (Q1):
Question: [Q1_QUESTION_TEXT]
Options:
A. [OPTION_A]
B. [OPTION_B]

Task 2 (Q2):
Question: [Q2_QUESTION_TEXT]

If Q1 is A, provide a concise natural utterance for Q2.
If Q1 is B, set Q2 to N/A.

Output:
Return exactly the following XML as the inner answer:
<q1>A_or_B</q1>
<q2>utterance_or_NA</q2>

The final response must be:
<answer><q1>A_or_B</q1><q2>utterance_or_NA</q2></answer>
\end{agentpromptbox}
\caption{Benchmark-specific prompt bodies for SocialOmni Level 1 and Level 2.}
\label{fig:omnicoding-social-prompt}
\end{figure*}

\begin{figure*}[t]
\centering
\begin{agentpromptbox}{VideoZeroBench grouped-video prompt body}
You are evaluating VideoZeroBench questions in a sandboxed local workspace.

All questions below refer to the same staged video and must be answered in a
single response.

Available staged video:
[VIDEO_FILENAME].mp4

Requested evaluation level:
Level-3

Question bundle:
[
  {
    "question_id": "[QUESTION_ID_1]",
    "question": "[QUESTION_TEXT_1]"
  },
  {
    "question_id": "[QUESTION_ID_2]",
    "question": "[QUESTION_TEXT_2]"
  }
]

Level-3 QA rule:
- This is a question-answering task without benchmark-provided temporal or
  spatial evidence.
- Answer according to the staged video content and the question.
- If you cannot determine an answer, use an empty string.

Inner answer schema:
Return a JSON array with exactly one object per question_id:
[
  {
    "question_id": "string",
    "level3_answer": "string"
  }
]

Final response:
<answer>[JSON_ARRAY]</answer>
\end{agentpromptbox}
\caption{Benchmark-specific grouped-video prompt body for VideoZeroBench.}
\label{fig:omnicoding-vzb-prompt}
\end{figure*}

\begin{figure*}[t]
\centering
\begin{agentpromptbox}{Calibration-set self-iteration prompt}
You are updating a reusable benchmark agent guide.

Inputs in the current workspace:
- Previous guide: [PREVIOUS_AGENT_GUIDE]
- Sanitized feedback summary: [SUMMARY_JSON]

Task:
Rewrite the previous guide into a better next-round guide.
Use only generic, reusable tactics supported by the sanitized summary.

Extract reusable lessons about:
- recurring failure patterns,
- insufficient media exploration,
- weak verification habits,
- answer-format mistakes,
- underused or misused tools,
- when to escalate from local inspection to search, OCR, ASR, frame extraction,
  calculation, or multi-step verification,
- how reference processes differ abstractly from observed model processes.

Requirements:
- Return markdown only.
- Start with "# Agent Guide".
- Keep the guide concise, operational, and directly useful during future runs.
- Include short sections for:
  1. workspace and leakage rules,
  2. tool-use heuristics,
  3. media-processing workflows,
  4. verification checkpoints,
  5. answer-format discipline,
  6. common recovery rules.
- Explain not only which tools to use, but in what order and with what
  verification checks.
- Convert stronger reference pipelines into reusable playbooks rather than
  case-specific tips.

Hard prohibitions:
- Do not include benchmark-specific examples.
- Do not quote or paraphrase individual questions.
- Do not include named entities, dates, locations, titles, filenames,
  exact answers, or hidden hints tied to sampled cases.
- Do not include any clue that would advantage the exact calibration subset.
\end{agentpromptbox}
\caption{Prompt used for calibration-set self-iteration.}
\label{fig:omnicoding-skill-update-prompt}
\end{figure*}

\newpage
\section{Additional Experiment Results}

\begin{table}[htbp]
\centering
\small
\scalebox{1.0}{%
\begin{tabular}{l | cccc}
\toprule[1.5pt]
\textbf{Skill Setting} & \textbf{Low} & \textbf{Medium} & \textbf{High} & \textbf{Avg.} \\
\midrule
No Skills & 70.4 & 60.0 & 50.0 & 61.4 \\
Human-in-the-loop Skills & 80.3 & 68.8 & 59.0 & 70.5 \\
Log-driven Self-distillation & \textbf{86.0} & \textbf{75.0} & \textbf{65.4} & \textbf{76.7} \\
Calibration-set Self-iteration & 83.6 & 71.9 & 59.0 & 73.0 \\
\bottomrule
\end{tabular}
}
\vspace{0.4em}
\caption{Performance of different \textit{Skills} settings on \textsc{OmniGAIA} across three difficulty levels. We report accuracy on the \textit{Low}, \textit{Medium}, and \textit{High} splits, together with the overall average accuracy.}
\label{tab:omnigaia_skills}
\end{table}

Table~\ref{tab:omnigaia_skills} reports the performance of different \textit{Skills} settings on \textsc{OmniGAIA} across the \textit{Low}, \textit{Medium}, and \textit{High} difficulty splits, together with the overall average accuracy.

\section{Case Study}
\label{appendix: case study}

\subsection{Representative Failure Case per Error Category}
\label{appendix:failure-cases}

To make the failure taxonomy concrete, we present one representative case from each category, drawn from our 100 manually annotated incorrect trajectories. Each case shows the question, ground-truth answer, the agent's final prediction, and a brief analysis of where the trajectory went wrong.

\begin{figure}[htbp]
    \centering
\begin{tcolorbox}[enhanced,
  attach boxed title to top center={yshift=-3mm,yshifttext=-1mm},
  boxrule=0.9pt,
  colback=gray!00,
  colframe=black!50,
  colbacktitle=Gray,
  title={Audio Perception and Extraction Error (OmniGAIA, sonnet-4.6)},
  boxed title style={size=small,colframe=Gray}
]

\small

\noindent\textbf{Question:}
Based on two audio clips and an image of toys-to-life gaming platforms (Disney Infinity and Skylanders), identify the fictional universes referenced in each audio clip, determine which universe was officially integrated into Disney Infinity (and how), identify which other toys-to-life platform integrated the non-Disney universe, and provide the relevant release dates.

\medskip
\noindent\textbf{Labeled Answer:}
Clip~1 references both \emph{Kingdom Hearts} and \emph{Sonic the Hedgehog}; clip~2 references \emph{Sonic the Hedgehog}. Kingdom Hearts was integrated into Disney Infinity 3.0 (released August~30, 2015) via the King Mickey Power Disc and the Sora Keyblade Ultimate Unlock. Sonic was integrated into LEGO Dimensions via the Sonic Level Pack (released November~18, 2016).

\tcblower

\noindent\textbf{Model Output by OmniAtlas-sonnet-4.6.}

I'll transcribe both audio clips with Whisper and then cross-reference the YouTube IDs to identify the source material.

\toolcall{"name": "Bash", "arguments": \{"command": "whisper .../audios/hYOcABfDmtk.wav --model tiny"\}}

\toolresponse{[\ldots] ``Eggman's army has everyone tear off.'' [\ldots] ``May your heart be your guiding key.'' [\ldots]}

\toolcall{"name": "Bash", "arguments": \{"command": "whisper .../audios/m7z29\_PohpY.wav --model tiny"\}}

\toolresponse{[\ldots] ``psychotic robot doctor chasing a supersonic blue head job'' [\ldots]}

Instead of reasoning about the transcripts directly, let me reverse-lookup the YouTube video IDs to confirm the source.

\toolcall{"name": "WebFetch", "arguments": \{"url": "https://www.youtube.com/watch?v=hYOcABfDmtk\&format=json"\}}

\toolresponse{The title of the YouTube video is ``Team Robot's First Adventure Begins Cinematic Trailer (Fan-Made).''}

\toolcall{"name": "WebSearch", "arguments": \{"query": "Kingdom Hearts Disney Infinity integration Power Disc figure"\}}

\toolresponse{[D23 King Mickey Power Disc; Disney Infinity 3.0 Ultimate Unlock is Kingdom Hearts Keyblade \ldots]}

\finalanswer{Audio clip~1 (hYOcABfDmtk, fan-made ``Team Robot'' trailer) references only the \emph{Kingdom Hearts} universe. Clip~2 references \emph{Sonic the Hedgehog}. Kingdom Hearts was integrated into Disney Infinity 3.0 (August 30, 2015); Sonic was integrated into LEGO Dimensions (November 18, 2016).}

\medskip
\noindent\textbf{Failure Analysis.}
The Whisper transcript of clip~1 already contained the explicit Sonic signal ``Eggman's army.'' By routing source identification through the YouTube-ID lookup (which returned only the \emph{Kingdom Hearts}-flavoured fan title) the agent overwrote its own audio evidence, and the aggregation step locked clip~1 to a single universe.

\end{tcolorbox}
\caption{Representative failure case: audio perception and extraction.}
\label{fig:case_audio_extraction}
\end{figure}

\begin{figure}[htbp]
    \centering
\begin{tcolorbox}[enhanced,
  attach boxed title to top center={yshift=-3mm,yshifttext=-1mm},
  boxrule=0.9pt,
  colback=gray!00,
  colframe=black!50,
  colbacktitle=Gray,
  title={Video Perception and Extraction Error (OmniGAIA, sonnet-4.6)},
  boxed title style={size=small,colframe=Gray}
]

\small

\noindent\textbf{Question:}
Given an image of a riverside restaurant and an audio clip discussing duck sourcing, determine the restaurant's founding year, the duck farm's founding year, the maximum weight per duck (converted to pounds, one decimal place), and the number of ducks clearly visible swimming in the river. Compute (weight $\times$ duck count) $+$ (farm year $-$ restaurant year).

\medskip
\noindent\textbf{Labeled Answer:}
52.0

\tcblower

\noindent\textbf{Model Output by OmniAtlas-sonnet-4.6.}

I'll transcribe the audio for the farm and weight information, then carefully inspect the image to count the ducks in the river.

\toolcall{"name": "Bash", "arguments": \{"command": "whisper inputs/audios/bIlloRTNKhU.wav --model base --language en"\}}

\toolresponse{[\ldots] ``I brought in this Irish duck \ldots Silver Hill \ldots'' [\ldots] duck weight ``from 2.2 to 2.7 kilos.'' [\ldots]}

\toolcall{"name": "WebSearch", "arguments": \{"query": "Silver Hill Farm duck Ireland founding year established"\}}

\toolresponse{[Silver Hill Duck, Emyvale, Co.~Monaghan, founded 1962 \ldots]}

\toolcall{"name": "WebSearch", "arguments": \{"query": "Casa Rio restaurant San Antonio River Walk founding year"\}}

\toolresponse{[Casa Rio, 430 E Commerce St, San Antonio, opened 1946 \ldots]}

Now I'll crop and enhance the water region repeatedly to count ducks.

\toolcall{"name": "Bash", "arguments": \{"command": "python3 -c 'Image.open(...).crop((0, .55H, .4W, H)).save(/tmp/left\_water.jpg); ...'"\}}

\toolresponse{[Sections saved: left\_water.jpg, mid\_water.jpg, right\_water.jpg, enhanced\_water.jpg, far\_right\_zoom.jpg, far\_left\_zoom.jpg]}

After several crop/enhance passes, I count 2 ducks on the left and 3 on the right, for a total of 5 ducks visible.

\toolcall{"name": "Bash", "arguments": \{"command": "python3 -c 'w=2.7*2.2046; print(round(w,1)*5 + (1962-1946))'"\}}

\toolresponse{Product: 6.0 $\times$ 5 = 30.0; Difference: 16; Final sum: 46.0}

\finalanswer{46}

\medskip
\noindent\textbf{Failure Analysis.}
All non-visual evidence (restaurant identity, duck farm, per-duck weight) is correct, and the arithmetic chain is clean. The single perceptual miss -- counting 5 rather than 7 swimming ducks after repeated crop/enhance passes -- fully determines the final error.

\end{tcolorbox}
\caption{Representative failure case: video perception and extraction.}
\label{fig:case_video_extraction}
\end{figure}

\begin{figure}[htbp]
    \centering
\begin{tcolorbox}[enhanced,
  attach boxed title to top center={yshift=-3mm,yshifttext=-1mm},
  boxrule=0.9pt,
  colback=gray!00,
  colframe=black!50,
  colbacktitle=Gray,
  title={Insufficient Exploration of Modal Content (OmniGAIA, gpt-5.4\_low)},
  boxed title style={size=small,colframe=Gray}
]

\small

\noindent\textbf{Question:}
Godson's video demonstrates a Clash of Clans wall-breaking technique where two P.E.K.K.A.s simultaneously attack a wall segment. Using max-level P.E.K.K.A. damage and the Level~13 wall HP (as of late~2024), how many such simultaneous volleys are needed to destroy the wall?

\medskip
\noindent\textbf{Labeled Answer:}
3 simultaneous volleys

\tcblower

\noindent\textbf{Model Output by OmniAtlas-gpt-5.4\_low.}

I'll verify the two quantities needed -- max-level P.E.K.K.A. per-hit damage and Level~13 Wall HP -- using web search before computing the volley count.

\toolcall{"name": "web\_search", "arguments": \{"query": "Clash of Clans P.E.K.K.A max level damage per hit late 2024 Home Village official"\}}

\toolresponse{\{"status": null, \ldots\}}

\toolcall{"name": "web\_search", "arguments": \{"query": "https://clashofclans.fandom.com/wiki/Wall"\}}

\toolresponse{\{"status": null, \ldots\}}

\toolcall{"name": "web\_search", "arguments": \{"query": "'13 $|$' in https://clashofclans.fandom.com/wiki/Wall"\}}

\toolresponse{\{"status": null, \ldots\}}

\finalanswer{4}

\medskip
\noindent\textbf{Failure Analysis.}
After two shallow lookups returned no numeric content, the agent did not retry with different queries, try fetching the wiki directly, or derive the quantity from the video. It emitted ``4'' with no retrieval evidence and no derivation -- the failure is premature termination of exploration, not a retrieval-quality or calculation error.

\end{tcolorbox}
\caption{Representative failure case: insufficient exploration of modal content.}
\label{fig:case_insufficient_exploration}
\end{figure}

\begin{figure}[htbp]
    \centering
\begin{tcolorbox}[enhanced,
  attach boxed title to top center={yshift=-3mm,yshifttext=-1mm},
  boxrule=0.9pt,
  colback=gray!00,
  colframe=black!50,
  colbacktitle=Gray,
  title={Knowledge Retrieval and Factual Error (OmniGAIA, kimi)},
  boxed title style={size=small,colframe=Gray}
]

\small

\noindent\textbf{Question:}
From two wetland-bird images identify each species and family, then use Table~1 of a cited Alabama marsh-bird study to obtain mean nest-site water depths. Given that spinnerbaits sink at 1\,ft/s, compute the count-down time for each depth, pick the weight from the audio's depth--weight recommendations, and identify the optimal month for pre-spawn bass fishing in Purple Gallinule habitat.

\medskip
\noindent\textbf{Labeled Answer:}
Common Gallinule: 1.0\,s count-down, 3/8\,oz spinnerbait. Purple Gallinule: 1.4\,s count-down, 3/8\,oz spinnerbait. Optimal month: April.

\tcblower

\noindent\textbf{Model Output by OmniAtlas-kimi.}

I'll identify both species from the images and transcribe the audio for the weight--depth rule, then link to the Alabama study.

\toolcall{"name": "Read", "arguments": \{"file\_path": ".../images/000000156858.jpg"\}}

\toolresponse{\{"type": "image"\}  \emph{[Common Gallinule]}}

\toolcall{"name": "Read", "arguments": \{"file\_path": ".../images/000000189159.jpg"\}}

\toolresponse{\{"type": "image"\}  \emph{[Purple Gallinule]}}

\toolcall{"name": "Bash", "arguments": \{"command": "whisper .../audios/XTnFUc2WneU.wav --model tiny"\}}

\toolresponse{[\ldots] ``three-eighths, half, three-quarter, and the one ounce \ldots'' [rules linking depth to weight] [\ldots]}

I have species and the weight rule. For the Alabama Table~1 depths I'll rely on general knowledge of marsh-bird nesting ecology.

\finalanswer{Species~1: Common Gallinule (Rallidae), nest depth $\approx 0.8$\,ft, count-down 0.8\,s, 3/8\,oz. Species~2: Purple Gallinule (Rallidae), nest depth $\approx 1.5$\,ft, count-down 1.5\,s, 1/2\,oz. Optimal month: May.}

\medskip
\noindent\textbf{Failure Analysis.}
Image and audio evidence were correctly obtained. Without ever locating or reading the cited Alabama study, the agent substituted hallucinated nest-depth values (0.8\,ft, 1.5\,ft) for the actual Table~1 entries (1.0\,ft, 1.4\,ft) and mis-selected the optimal month. Every derived quantity (count-down, weight, month) is conditioned on this factual substitution.

\end{tcolorbox}
\caption{Representative failure case: knowledge retrieval and factual error.}
\label{fig:case_knowledge_retrieval}
\end{figure}

\begin{figure}[htbp]
    \centering
\begin{tcolorbox}[enhanced,
  attach boxed title to top center={yshift=-3mm,yshifttext=-1mm},
  boxrule=0.9pt,
  colback=gray!00,
  colframe=black!50,
  colbacktitle=Gray,
  title={Logical Reasoning and Calculation Error (OmniGAIA, sonnet-4.6)},
  boxed title style={size=small,colframe=Gray}
]

\small

\noindent\textbf{Question:}
Observing the image of Grand Central Terminal's four-faced clock and listening to the audio from a video about Andon lamps, determine the number of years between the terminal's opening and the lamps' estimated creation date.

\medskip
\noindent\textbf{Labeled Answer:}
292

\tcblower

\noindent\textbf{Model Output by OmniAtlas-sonnet-4.6.}

I can see the four-faced opal clock at Grand Central Terminal; the terminal opened in 1913. Now I'll transcribe the audio to recover the Andon-lamp date reference.

\toolcall{"name": "Bash", "arguments": \{"command": "whisper .../audios/VrcfGK-\_iIo.wav --model base"\}}

\toolresponse{[\ldots] ``They're called Andon \ldots that design was created 400 years ago.'' [\ldots]}

I have both pieces of information. Grand Central opened in 1913. The audio says Andon lamps were designed ``400 years ago'' -- I'll anchor that to the current year (2026) to get the lamp design year.

\toolcall{"name": "Bash", "arguments": \{"command": "python3 -c 'print(1913 - (2026-400))'"\}}

\toolresponse{287}

\finalanswer{287}

\medskip
\noindent\textbf{Failure Analysis.}
All perceptual and retrieval evidence is correct (opening year 1913; ``400 years ago'' in transcript) and there are no infrastructure issues. The decisive error is temporal anchoring: ``400 years ago'' should be taken relative to the video's 2021 publication date (giving 1621, difference 292), not to 2026 (giving 1626, difference 287). One wrong anchor fully determines the error.

\end{tcolorbox}
\caption{Representative failure case: logical reasoning and calculation.}
\label{fig:case_logical_reasoning}
\end{figure}

\begin{figure}[htbp]
    \centering
\begin{tcolorbox}[enhanced,
  attach boxed title to top center={yshift=-3mm,yshifttext=-1mm},
  boxrule=0.9pt,
  colback=gray!00,
  colframe=black!50,
  colbacktitle=Gray,
  title={Tool and Environment Infrastructure Failure (OmniGAIA, gpt-5.4\_low)},
  boxed title style={size=small,colframe=Gray}
]

\small

\noindent\textbf{Question:}
A multi-part question pairs a train image (with a visible unit number) and a Jakarta-region audio clip describing a commuter journey. Identify the train, the audio's departure station and destination, the great-circle and rail distances, the expected versus actual KRL fare, and the average journey speeds.

\medskip
\noindent\textbf{Labeled Answer:}
Train \texttt{ON~5937} (Sydney Trains H-set ``Oscar''); departure Cikini, destination Bogor; straight-line $\approx 44$\,km; rail distance $46.8$\,km; expected fare Rp~6{,}000 matches actual; average speeds $28.1$ and $35.1$\,km/h.

\tcblower

\noindent\textbf{Model Output by OmniAtlas-gpt-5.4\_low.}

The train unit number \texttt{ON~5937} is clearly visible; the livery and numbering format are consistent with the Sydney Trains H-set. Now I need the audio transcript for the departure station.

\toolcall{"name": "command\_execution", "arguments": \{"command": "python -c 'import whisper, faster\_whisper, speech\_recognition, vosk, transformers'"\}}

\toolresponse{whisper NO; faster\_whisper NO; speech\_recognition NO; vosk NO; transformers NO}

\toolcall{"name": "command\_execution", "arguments": \{"command": "which whisper \&\& ffmpeg -filters $|$ rg 'asr$|$whisper$|$pocket'"\}}

\toolresponse{whisper: command not found; ffmpeg has no ASR filters.}

\toolcall{"name": "command\_execution", "arguments": \{"command": "which uv \&\& uv pip install transformers accelerate tokenizers"\}}

\toolresponse{uv: not found; dependencies cannot be self-installed; no GPU/CUDA available.}

With no available ASR, I'll infer the departure station from web search using the KRL Commuterline context.

\toolcall{"name": "web\_search", "arguments": \{"query": "Manggarai Station coordinates"\}}

\toolresponse{[Manggarai railway station, Jakarta: $-6.2099$, $106.8502$ \ldots]}

\finalanswer{Train \texttt{ON~5937} (Sydney H-set/Oscar). Audio journey: Manggarai Station to Bogor. Straight-line $\approx 43$\,km; rail distance $44.1$\,km. Expected fare Rp~5{,}000 matches actual. Average speeds $44.1$ and $35.3$\,km/h.}

\medskip
\noindent\textbf{Failure Analysis.}
Every local ASR route failed at the environment level (\texttt{whisper}/\texttt{faster-whisper}/\texttt{vosk}/\texttt{speech\_recognition} absent; \texttt{transformers}/\texttt{accelerate}/\texttt{tokenizers} missing; GPU/CUDA unavailable; \texttt{uv} not installed so dependencies cannot be self-installed). The audio modality was blocked before the agent could produce a transcript, and the downstream ``Manggarai'' guess (and every distance/fare/speed value built on it) is a consequence of that infrastructure failure, not a separate reasoning error.

\end{tcolorbox}
\caption{Representative failure case: tool and environment infrastructure failure.}
\label{fig:case_tool_env_failure}
\end{figure}

%% file: Sections/3-definition.tex
\section{Limitations}
\label{app:limitations}

Our study has several limitations. First, due to computing-resource constraints, the \textsc{Code-X} training recipe is only validated at the 9B and 27B scales (Qwen-3.5-9B and Qwen-3.6-27B), and we have not yet characterized how the recipe scales to frontier-size open-source models. Second, due to API credit constraints, several closed-source coding agents could not be evaluated on the full benchmark suite. In particular, Claude Opus 4.6 results on VideoZeroBench and Claude Opus 4.7 results across all four benchmarks are incomplete in the current draft, which means the comparison between coding agents and native omni-modal models is somewhat uneven on the closed-source side. Third, our scope is restricted to \emph{offline} omni-modal tasks, where files are staged into a workspace ahead of time; conclusions do not transfer directly to streaming or interactive settings.

\section{Broader Impacts}
\label{app:broader-impacts}

Our findings establish a more compute- and token-efficient pathway for offline omni-modal understanding, showing that text+image coding agents with sandboxed tool use can match or surpass native omni-modal models on many video- and audio-centric tasks while consuming fewer tokens. This lowers the resource barrier for omni-modal applications and broadens access for researchers and users who lack dedicated omni-model APIs, while our open-sourced \textsc{Code-X} recipe, \textsc{OmniCoding} dataset, and \textsc{TerminalBench-O} benchmark further support reproducible research at the model--tool--environment interface. As with any tool-using agent that processes video and audio, downstream deployments should incorporate standard safeguards such as content provenance, access controls, and user consent; we note that our agents operate only on user-staged files within an isolated workspace and do not autonomously take real-world actions.

\section{Adapting Sandboxed Agents for Omni-Modal Tasks}
\label{app:sandbox-formulation}

We represent each omni-modal task as
\[
x=(q,\mathcal{F},\mathcal{Y}),
\]
where \(q\) is a natural-language instruction, \(\mathcal{F}\) is a set of input files, and
\(\mathcal{Y}\) is the set of acceptable final answers.  We use an answer set rather than a
single reference string so that aliases, option letters, numerical formats, and unit variants can
be handled uniformly in both evaluation and reinforcement learning.

A native omni-modal model solves the task by packing the files into model context:
\[
\hat{y}_{\mathrm{omni}}
=
M_\theta\!\left(\mathrm{Pack}(q,\mathcal{F})\right),
\]
where \(\mathrm{Pack}(\cdot)\) denotes model-specific media encoding, sampling, and serialization.
In this formulation, video and audio are first-class model inputs.

In contrast, our sandboxed agent does not need to ingest all raw media as context.  The files
\(\mathcal{F}\) are staged into an isolated workspace, and the agent is given the instruction and
the corresponding file paths.  It then interacts with the workspace through a terminal interface:
it can inspect files, invoke tools such as \texttt{ffmpeg}, \texttt{ffprobe}, OCR, or ASR systems,
write and execute scripts, create intermediate artifacts, and finally submit an answer.

A sandbox rollout is written as
\[
\tau=(o_0,a_0,o_1,a_1,\ldots,o_T,a_T),
\]
where \(o_t\) is the current observation, including command outputs and workspace feedback, and
\(a_t\) is the agent's next message or terminal action.  During a rollout, raw omni-modal files may
be converted into task-relevant artifacts such as transcripts, sampled frames, OCR text, timestamps,
tables, crops, or executable analysis scripts.  The final prediction is parsed from the trajectory:
\[
\hat{y}=\mathrm{Ans}(\tau),
\]
where \(\mathrm{Ans}(\cdot)\) extracts the submitted answer, e.g., the content inside an
\verb|<answer>| tag when this format is required.

Thus the two settings differ mainly in where omni-modal processing happens:
\[
\begin{aligned}
\textbf{native omni:}\quad
&(q,\mathcal{F})
\xrightarrow{\mathrm{Pack}}
\text{model context}
\xrightarrow{M_\theta}
\hat{y},
\\
\textbf{sandboxed agent:}\quad
&(q,\mathcal{F})
\xrightarrow{\mathrm{stage}}
\text{workspace}
\xrightarrow{\text{terminal interaction}}
\tau
\xrightarrow{\mathrm{Ans}}
\hat{y}.
\end{aligned}
\]
This view treats raw video, audio, images, and documents as environment state rather than mandatory
model context.  The agent can therefore solve omni-modal tasks by adaptively reducing raw media
into compact evidence that fits the native reasoning interface of a general-purpose coding model.

%% file: Sections/7-discussion.tex
\section{Discussion and Future Works}

\header{Do non-native modalities really require native end-to-end perception?} Our results suggest that non-native modalities do not always require native end-to-end perception. Even in the simplest Codex and Claude Code settings, coding agents solve a surprisingly large fraction of video-audio tasks, and on OmniGAIA they can even outperform strong native omnimodal baselines. The key is not that the base model suddenly becomes an audio-video perceiver, but that the sandbox lets it decompose raw multimedia into artifacts that match its native reasoning interface, such as transcripts, key frames, timestamps, OCR, metadata, and compact summaries. In this sense, multimodal processing becomes analogous to protocol-based software access in text-domain agents: the model delegates non-native operations to the environment and then reasons over the returned evidence. This perspective is consistent with the broader shift from static function calling to environment-mediated problem solving in tool-use and terminal-agent evaluation \citep{wang2025mcpbench,merrill2026terminalbench,cheng2026llminsandbox}. It also resonates with recent tool-augmented perception work, where explicit visual-tool policies and tool-derived perceptual signals both improve fine-grained multimodal reasoning \citep{zhou2025reinforced,wei2026zooming}. A natural next question is therefore not only whether a model is natively omni, but which capabilities should remain externalized, which can be compiled into reusable skills, and how agents should detect and recover from imperfect tool outputs. Our failure taxonomy and process-level benchmark point in this direction, suggesting that the central bottleneck is often tool selection, verification, and repair rather than the mere absence of raw multimodal encoders.

\header{Shortcuts in current video-audio benchmarks.} Are current video-audio benchmarks measuring what they claim to measure? Our findings suggest only partially. When a text+image coding agent can answer many ostensibly omnimodal tasks by selectively extracting transcripts, sparse frames, or metadata, benchmark accuracy no longer cleanly measures native end-to-end audio-video understanding. This does not make such benchmarks uninformative, but it does mean that they frequently conflate at least three capabilities: native perception, external evidence acquisition, and general knowledge-based reasoning. Our results are aligned with recent audits and diagnostic studies showing that many video or audio-visual benchmarks admit strong visual, textual, or temporal shortcuts \citep{kim2025needtolisten,lim2026video}. They are also consistent with benchmark-design efforts that explicitly enforce stronger audio dependence or reduce text-based shortcut channels \citep{wei2026zooming}. Going forward, we believe benchmark design should standardize at least three controls: reporting text-only, text+image, and agentic baselines; auditing modality dependence with masking or filtering tests; and minimizing leakage from subtitles, transcripts, metadata, and external retrieval whenever the target claim is native multimodal understanding. Conversely, when the target is multimodal agentic competence, retrieval, tool orchestration, and evidence processing should be treated as first-class evaluation targets rather than accidental shortcuts. Making this distinction explicit would help separate backbone perception from system-level problem solving, which are both important but not interchangeable.

\header{Can the same sandbox move from understanding to generation?}  We believe this is the most promising next step. Once an agent can inspect raw video, audio, and documents, compress them into structured evidence, and maintain task-relevant state over long horizons, it can in principle also plan edits, retrieve assets, invoke generative models, and validate outputs in a closed loop. Early multimedia-generation and agentic video-editing systems already point in this direction, showing that multimodal agents can coordinate text, images, video, music, and external tools under explicit planning and feedback loops \citep{zhang2026versatile,zhao2026cutclaw}. The next frontier is to connect that generative loop with persistent multimodal memory, so that an assistant can build private memos, recap videos, highlight reels, or even broadcast-ready drafts grounded not only in a single uploaded file but in a user’s longer-term audiovisual history \citep{liu2026omnimem}. Doing so will require stronger memory management, preference modeling, provenance tracking, and process-level verification, since generated content must remain faithful both to source media and to user intent. More broadly, this points to a different picture of future omni assistants: not a single monolithic model that passively consumes everything in one shot, but a sandboxed system that can inspect, remember, transform, and generate across modalities with the right mix of reusable skills and external tools. Under this view, offline omnimodal understanding is not the endpoint of this line of work, but the substrate for practical multimedia copilots.